\RequirePackage{fix-cm}
\documentclass[twocolumn]{svjour3}          % twocolumn

\smartqed  % flush right qed marks, e.g. at end of proof
\usepackage{graphicx}
\usepackage{natbib}
\usepackage{multirow}
\usepackage{amsmath}
\usepackage{amssymb}
\usepackage{bbm}
\usepackage{bm}
\usepackage{booktabs}
\usepackage{array} % for extended column definitions, e.g. >{\em}c
\usepackage{blindtext}
\usepackage{comment}
\usepackage{booktabs}
\usepackage{multirow}
\usepackage{multicol}
\usepackage[mathscr]{euscript}
\usepackage[dvipsnames]{xcolor}
\usepackage{colortbl}
\usepackage{makecell}
\usepackage{subcaption}
\usepackage[symbol]{footmisc}

\usepackage{pifont}

\usepackage{marvosym}
\usepackage{ifsym}
\usepackage{xspace}

\newcommand{\etal}{\textit{et al.}\xspace}

\usepackage[dvipsnames]{xcolor}
\usepackage{color, colortbl}
\definecolor{citecolor}{HTML}{0071bc}
\definecolor{tabhighlight}{HTML}{e5e5e5}
\usepackage[colorlinks,citecolor=citecolor]{hyperref}
\makeatletter
\renewcommand\paragraph{
  \@startsection{paragraph} % name
  {4} % level
  {\z@} % indent
  {.5em \@plus1ex \@minus.2ex} % beforeskip
  {-.5em} % afterskip
  {\normalfont\normalsize\bfseries} % style
}
\makeatother
\begin{document}
\sloppy

\title{Fg-T2M++: LLMs-Augmented Fine-Grained Text Driven Human Motion Generation 
}

%\titlerunning{Short form of title}        % if too long for running head

\author{Yin Wang \and
        Mu Li \and
        Jiapeng Liu \and
        Zhiying Leng \and
        Frederick W. B. Li \and
        Ziyao Zhang \and
        Xiaohui Liang \Letter
}

\authorrunning{Yin Wang et al.} % if too long for running head

\institute{
        Yin Wang \at
              State Key Laboratory of Virtual Reality Technology and Systems, Beihang University, Beijing, China \\
              \email{wang\_yin@buaa.edu.cn}      
        \and
            Mu Li \at
              State Key Laboratory of Virtual Reality Technology and Systems, Beihang University, Beijing, China \\
              \email{limu@buaa.edu.cn}
          \and
          Jiapeng Liu \at
              State Key Laboratory of Virtual Reality Technology and Systems, Beihang University, Beijing, China \\
              \email{zy2306414@buaa.edu.cn}             
          \and
          Zhiying Leng \at
              State Key Laboratory of Virtual Reality Technology and Systems, Beihang University, Beijing, China \\
              \email{zhiyingleng@buaa.edu.cn}            
          \and
          Frederick W. B. Li \at
              Department of Computer Science, University of Durham, U.K \\
              \email{frederick.li@durham.ac.uk}        
          \and
                Ziyao Zhang \at
              State Key Laboratory of Virtual Reality Technology and Systems, Beihang University, Beijing, China \\
              \email{20373042@buaa.edu.cn}                      
          \and
          Xiaohui Liang (\Letter Corresponding author)\at
              State Key Laboratory of Virtual Reality Technology and Systems, Beihang University, Beijing, China \\
              Zhongguancun Laboratory, Beijing, China\\
              \email{liang\_xiaohui@buaa.edu.cn}
}

\date{Received: date / Accepted: date}
% The correct dates will be entered by the editor

\maketitle

\begin{abstract}
We address the challenging problem of fine-grained text-driven human motion generation. Existing works generate imprecise motions that fail to accurately capture relationships specified in text due to: (1) lack of effective text parsing for detailed semantic cues regarding body parts, (2) not fully modeling linguistic structures between words to comprehend text comprehensively. To tackle these limitations, we propose a novel fine-grained framework Fg-T2M++ that consists of: (1) an \emph{LLMs semantic parsing module} to extract body part descriptions and semantics from text, (2) a \emph{hyperbolic text representation module} to encode relational information between text units by embedding the syntactic dependency graph into hyperbolic space, and (3) a \emph{multi-modal fusion module} to hierarchically fuse text and motion features. Extensive experiments on HumanML3D and KIT-ML datasets demonstrate that Fg-T2M++ outperforms SOTA methods, validating its ability to accurately generate motions adhering to comprehensive text semantics.


\end{abstract}

\section{Introduction}\label{sec1}

\begin{figure*}[t]
    \centering
    \includegraphics[width=0.825\linewidth]{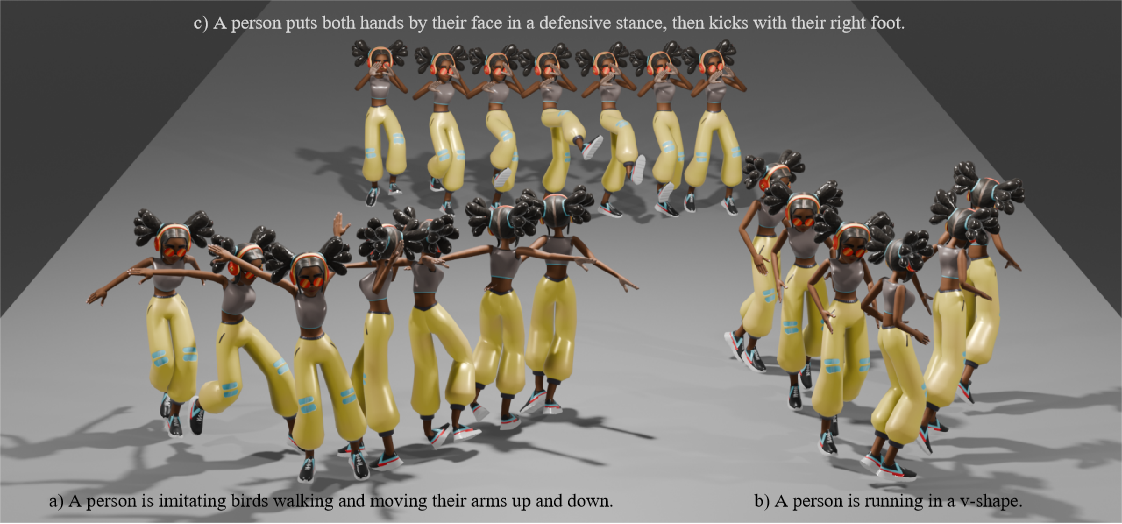}
    \caption{Our Fg-T2M++ excels in generating high-quality and diverse motion sequences, capturing fine-grained details embedded in the text prompts.}
    \label{teaser}
\end{figure*}

Human motion generation is a pivotal but challenging task in computer vision with applications in animation, AR/VR, gaming etc. While existing works utilize diverse multimodal inputs such as music \citep{kao2020temporally,li2021ai,ren2020self,starke2022deepphase,tseng2023edge}, motion categories \citep{guo2020action2motion,petrovich2021action,cervantes2022implicit,guo2022action2video} and trajectories \citep{karunratanakul2023guided,shafir2023human,wan2023tlcontrol}, collecting and annotating such data requires substantial expertise. Recently, text-driven human motion generation (T2M) has emerged as a promising research direction, which generates motions represented by 3D joint positions or rotations \citep{guo2022generating,petrovich2022temos,tevet2023human,chen2023executing,zhang2022motiondiffuse}. T2M holds potential for intuitive motion control through semantics-rich text inputs. However, key challenges remain in parsing intricate spatiotemporal relationships between body parts from descriptions and mapping diverse linguistic expressions to realistic motions. While promising an accessible interface, addressing these issues is non-trivial. Advances in T2M could unlock applications from animation to assistive technologies through natural guidance.

Three main approaches are proposed to address text-driven human motion generation: (1) Latent space alignment methods, such as JL2P \citep{ahuja2019language2pose} and TEMOS \citep{petrovich2022temos}, aim to learn a shared latent space between text and motion representations by directly integrating their embeddings. However, this integration can potentially lead to the loss of modality-specific information. (2) Conditional autoregressive models generate motion tokens sequentially, conditioned on previous tokens and text.
Pioneering works like TM2T \citep{guo2022tm2t} employ vector quantized VAEs to decode motion tokens from discrete representations learned from data, while T2M-GPT \citep{zhang2023generating} enhances this with techniques such as exponential moving average and code resetting for more natural generation. Despite their strengths in capturing temporal dependencies, these methods rely on unidirectional and sequential prediction, which can impact the quality of motion generation due to cumulative errors. (3) Conditional diffusion models, including MotionDiffuse \citep{zhang2022motiondiffuse} and MDM \citep{tevet2023human}, adopt diffusion frameworks to probabilistically map text to motion via denoising training objectives, achieving promising performance. The condition plays a crucial role in guiding the denoising process. However, current methods often lack refinement in handling these conditions. Firstly, the exploration of additional auxiliary information is insufficient because datasets like HumanML3D \citep{guo2022generating} and KIT-ML \citep{plappert2016kit} offer only coarse descriptions without fine-grained part-level annotations. Secondly, the extraction and fusion of conditional features are inadequate; current methods typically extract compact sentence representations from text, failing to fully utilize rich information within words. This limitation can result in generated motions deviating from the original text meaning when simply concatenating sentence-motion vectors.

Existing text-to-motion methods struggle with generating whole-body motions for unseen text, as full motion may lie outside the training distribution, yet individual body part motions still fall within it. We address this by hypothesizing decomposition of whole-body generation into combinable sub-joint motions of multiple parts facilitates easier modeling. Also, natural language semantically encodes actions through parts-of-speech and syntactically relates words through grammatical structures. To capture fine-grained linguistic details, we propose analyzing individual body part motions specified by words, considering their syntactic roles and relationships. Based on these insights, we introduce the novel framework Fg-T2M++ (Figure \ref{teaser}), leveraging part-level and word-level natural language descriptions to generate precise motions conditioned on text prompts. This approach decomposes generation and deeply analyzes textual details, aiming to overcome limitations in generating whole-body and fine-grained motions.

% Fg-T2M++ comprises three integrated components for fine-grained language-guided motion generation. First, \textcolor{blue}{to obtain more fine-grained information from text prompts}, our \emph{LLMs Semantic Parsing (LSP) module}, parsing text into detailed annotations of individual body part motions and their relationships, achieved through deep linguistic analysis of semantic roles between parts of speech (e.g. nouns, adjectives) and motions. This fine-grained parsing maps individual textual elements to joint movements, allowing understanding of more complex language than prior work relying on shallow encodings.  
% %
% Second, \textcolor{blue}{to extract fine-grained features from text prompts}, our \emph{Hyperbolic Text Representation (HTP) module}, constructs a dependency parse tree and embeds it in hyperbolic space. Hyperbolic geometry intrinsically preserves hierarchies with low distortion (\cite{yang2022hyperbolic}), allowing HTP to learn syntactic tree structure while modeling textual relationships with greater power than Euclidean linguistics models.
% %
% \textcolor{blue}{Third, to achieve a fine-grained fusion of multimodal information, our \emph{Multi-Modal Fusion (MMF) module}, hierarchically fuses the HTP output and LLMs features at both global and local levels.} This multi-scale fusion of syntactic and semantic information provides a comprehensive understanding of text-motion mappings not achieved by prior global or local modeling alone.

Fg-T2M++ comprises three integrated components for fine-grained language-guided motion generation, each serving distinct purposes. First, our \emph{LLMs Semantic Parsing (LSP) module} uses large language models to extract detailed semantic descriptions from text prompts. It parses the text into annotations of individual body part motions and their relationships through deep linguistic analysis of semantic roles between parts of speech (e.g., nouns, adjectives) and motions. This fine-grained parsing maps individual textual elements to joint movements, allowing for the understanding of complex language beyond prior methods that relied on shallow encodings.
Second, the \emph{Hyperbolic Text Representation (HTP) module} focuses on encoding the syntactic structure of text prompts by constructing a dependency parse tree and embedding it in hyperbolic space. Hyperbolic geometry intrinsically preserves hierarchies with low distortion \citep{yang2022hyperbolic}, enabling HTP to capture hierarchical relationships more effectively than Euclidean models.
Third, to achieve a fine-grained fusion of multimodal information, our \emph{Multi-Modal Fusion (MMF) module} hierarchically fuses outputs from the HTP and LSP modules at both global and local levels. It combines global and local features to learn comprehensive text-motion mappings. This multi-scale fusion of syntactic and semantic information provides a comprehensive understanding of text-motion mappings not achieved by prior global or local modeling alone.
By integrating linguistically-informed parsing, hyperbolic syntactic modeling, and hierarchical semantic fusion, Fg-T2M++ captures fine-grained text-motion correlations in a technically advanced yet concise manner compared to prior works.

While our previous work, Fg-T2M, in ICCV 2023 \citep{wang2023fg}, made progress in text-driven motion generation, it was incapable of addressing the novel research problems associated with capturing fine-grained motion details specified in text. This limitation arose from its coarse-grained modeling of syntactic relationships without a detailed analysis of text prompts. To address this, we propose Fg-T2M++ with novel technical contributions - LLMs Semantic Parsing to extract body part-level semantics from text, hyperbolic text representation module encoding hierarchical dependency graphs in hyperbolic space, and multi-modal fusion performing multi-level fusion within a conditional diffusion framework. This design further extends Fg-T2M's capabilities for fine-grained tasks. Extensive evaluation demonstrates the effectiveness of Fg-T2M++ over Fg-T2M, achieving a significantly lower FID of 0.135 versus 0.571 and MM-Dist of 2.696 versus 3.114 on KIT-ML, validating its ability to generate motions specified by richer textual details that Fg-T2M was technically unable to capture.
Our main contributions are:

\begin{itemize}
  \item We propose an LLMs Semantic Parsing module to parse text into fine-grained body part representations and detailed words semantics leveraging large language models.

  \item We introduce a Hyperbolic Text Representation module incorporating dependency parsing and hyperbolic graph convolution to embed syntactic trees in hyperbolic space, exploiting its advantages over Euclidean space.

  \item We present a Multi-Modal Fusion module performing hierarchical fusion of global and local text-motion relationships within a conditional diffusion framework through multiple denoising steps.

  \item We validate Fg-T2M++ on HumanML3D and KIT-ML datasets, demonstrating SOTA performance through metrics, and qualitative results revealing finer motion generation matching text.
\end{itemize}

\section{Related Work}\label{sec2}

\subsection{Text Driven Human Motion Generation}\label{sec2_1}

While latent space alignment works such as JL2P \citep{ahuja2019language2pose} and Ghosh et al. \citep{ghosh2021synthesis} achieved progress utilizing joint embeddings and hierarchical encoders capturing coarse relationships, as well as techniques like MotionCLIP \citep{tevet2022motionclip} that generates stylized motions by projecting into a shared space learned via CLIP \citep{radford2021learning}, TEMOS \citep{petrovich2022temos} combining motion and text VAEs, and temporal VAE \citep{guo2022generating} for sequence generation, their limitation is loss of fine-grained details when encoding independently. 

Autoregressive models such as TM2T \citep{guo2022tm2t}, which learns mutual mappings of motion and tokens via vector quantized VAEs, T2M-GPT \citep{zhang2023generating}, which enhances performance with EMA and code resetting, and AttT2M \citep{zhong2023attt2m} mapping to refined codes via body part attention, have achieved progress in representing motion as discrete tokens. However, the unidirectional nature of autoregressive models limits their ability to capture future context, affecting motion quality. Incorporating bidirectional dependencies could improve this but increase training and inference costs due to the additional computational complexity involved.

Recent diffusion-based models, such as MotionDiffuse \citep{zhang2022motiondiffuse}, MDM \citep{tevet2023human}, and FLAME \citep{kim2023flame}, have shown promising performance in T2M tasks by leveraging conditional diffusion to learn probabilistic text-motion mappings. Also, MLD \citep{chen2023executing} employs latent diffusion to enhance efficiency, while ReMoDiffuse \citep{zhang2023remodiffuse} incorporates sample retrieval for contextual understanding. However, they may suffer from a lack of fidelity in generating motions that precisely align with conditional inputs, particularly in capturing complex multi-modal relationships. 

% \textcolor{blue}{Existing methods perform relatively well on coarse-grained text, such as ``a person is walking." However, they still struggle with fine-grained text. For example, ``a person is walking with the right hand raising while stumbling to the left." The latent space alignment method loses a significant feature detail during feature projection, and may only process the common ``walking" motion. While autoregressive models, due to the unidirectional prediction feature, may overlook the influence of subsequent movements, such as ``stumbling," when generating the earlier movement of ``walking," leading to motion incoherence. Existing diffusion-based models, due to insufficient feature extraction and fusion between the denoising sequence and the text, may ignore the finer details, such as ``right hand raising'' or ``stumbling to the left".}

Existing methods perform relatively well on coarse-grained text, such as ``a person is walking.'' However, they struggle with fine-grained text that involves complex syntax-kinematic associations, like ``a person is walking with the right hand raising while stumbling to the left.'' Latent space alignment methods can lose significant feature details during feature projection, often processing only the common ``walking" motion. Autoregressive models, due to their unidirectional prediction nature, may overlook subsequent movements like ``stumbling'', leading to motion incoherence. Diffusion-based models face challenges in feature extraction and fusion between the denoising sequence and text, which can result in ignoring finer details such as ``right hand raising'' or ``stumbling to the left.''

Despite advances in text-to-motion generation, challenges remain regarding fine-grained modeling. The sparsity of current datasets limits learning precise textual cue-motion correspondences. Insufficient use of linguistic cues also restricts comprehending fine-grained semantics from prompts. Addressing these issues, we created detailed annotations of different body parts' actions and words explanations, enabling more intricate understanding of part-specific details. Furthermore, leveraging linguistic structures assists semantic parsing of prompts. This enables our model to generate human motions closely aligned with the semantic content of input text, exhibiting realistic movements.

\subsection{LLMs-Assisted Motion Generation}\label{sec2_2}

While large language models such as BERT \citep{devlin2018bert}, GPT-4 \citep{achiam2023gpt} and T5 \citep{raffel2020exploring} have demonstrated strong capabilities in language tasks as evidenced by their human-level performance in certain domains \citep{gilardi2023chatgpt}, their application to human motion generation has strengths and limitations. Recent works including ActionGPT \citep{kalakonda2023action}, SINC \citep{athanasiou2023sinc}, FineMoGen \citep{zhang2024finemogen}, and MotionGPT \citep{jiang2024motiongpt} have explored leveraging LLMs' language generation and zero-shot transfer abilities to enrich prompts, identify body parts, facilitate human-AI interaction, and support various motion-related tasks. However, directly generating coherent human motions from language remains challenging due to the complex grounding problem between language and bodily motion. Their suitability ultimately depends on how effectively language representations can condition low-dimensional movement sequences.

Previous works utilizing LLMs have achieved good results, yet challenges for improving fine-grained analysis and modeling persist. Urgently needed are datasets with precise, fine-grained text representations that are sensitive to subtle motion details. To address this, we introduce the use of LLMs for parsing text prompts at a fine level to obtain specific descriptions of individual body parts. We also provide detailed explanations of nouns, adjectives, and adverbs in sentences to address challenging vocabulary in complex texts. By training models with these fine-grained linguistic details regarding all body parts and words, we can generate high-fidelity, fine-level human motion sequences.

\section{Preliminaries}

\subsection{Diffusion Model} 

Our Fg-T2M++ model for fine-grained text-to-motion generation is based on the diffusion probabilistic framework. As described in prior work (\cite{ho2020denoising}), diffusion models comprise a forward noise injection process and reverse conditional generation process. 
In the forward process, a clean target motion sequence $x_0$ is gradually corrupted with added Gaussian noise to produce a simple Gaussian distribution. This defines an auto-encoding formulation.

Crucially, in the reverse process, noise is removed from the corrupted motion sequence $x_1,...,x_T$ in a conditional manner given natural language text $c$. We model this conditional generation as:
\begin{equation}
% p_{\theta}(x_{0:T}|c) = p(x_{T}|c) \prod_{t=1}^{T}p_{\theta}(x_{t-1}|x_{t},c)  
p_{\theta}(x_{0:T}|c) = p(x_{T}) \prod_{t=1}^{T}p_{\theta}(x_{t-1}|x_{t},c)
\end{equation}

By conditioning each step of the diffusion posterior on both the motion context and linguistic input $c$, our Fg-T2M++ model can progressively generate fine-grained target motions aligned precisely with the text description. This principled diffusion formulation underpins our ability to capture rich, detailed language-motion mappings.

\begin{figure*}[t]
    \centering
    \includegraphics[width=0.9\linewidth]{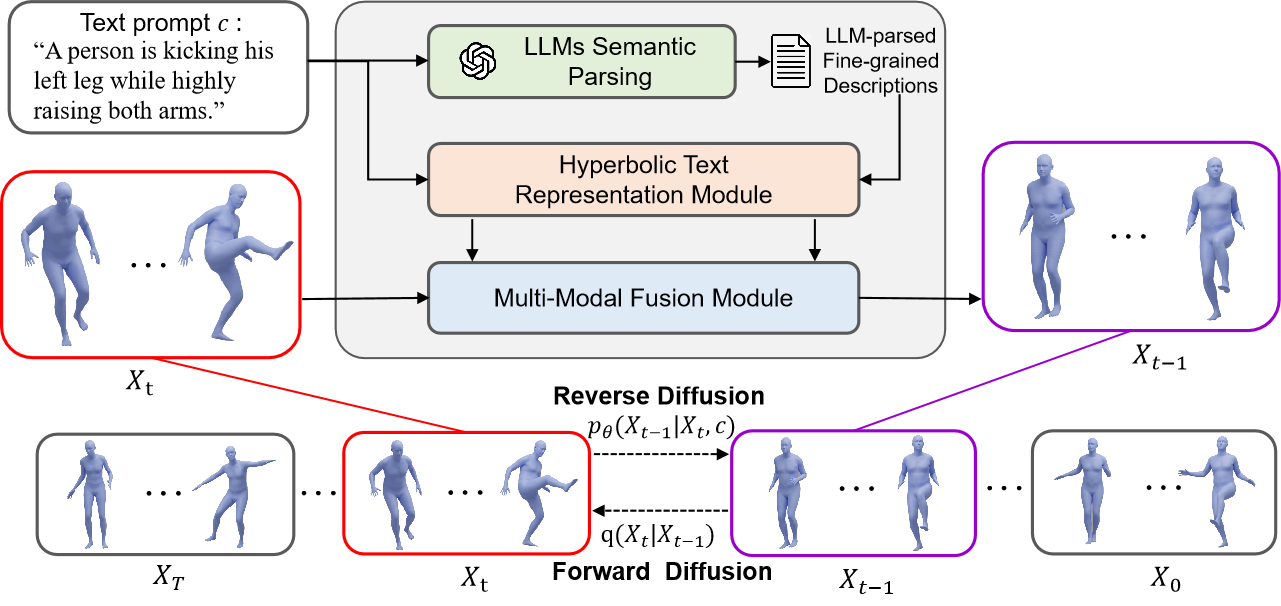}
    \caption{\textbf{Overview of Fg-T2M++}: Given a text prompt $c$, the reverse denoising process of the diffusion model starts from noisy motion data $X_T$ and produces clean motion data $X_0$. Initially, the text prompt undergoes LLMs semantic parsing to generate LLMs-parsed fine-grained descriptions. Then, both the text prompt and its parsed descriptions are input into the hyperbolic text representation module, which captures precise representations of text features. Finally, the noisy motion data $X_t$, along with the two fine-grained text features, are fed into the multi-modal fusion module to obtain the clean motion data $X_{t-1}$.}
    \label{pipeline}
\end{figure*}

\subsection{Hyperbolic Graph Convolution} \label{HGCN}

% Hyperbolic space $H^n$ is a non-Euclidean space of constant negative curvature that can be represented as an $n$-dimensional Riemannian manifold \citep{nickel2017poincare}. Among models like the Poincar\'e ball model, Lorentz model and Klein model, we utilize the Poincar\'e ball model $B_c^n=\{x\in\mathbb{R}^n||x||^2<r^2\}$ due to its geometric properties preservation \citep{nickel2017poincare}.

Hyperbolic space \(H^n\) is a non-Euclidean space with constant negative curvature, represented as an \(n\)-dimensional Riemannian manifold. Among models like the Poincaré ball, Lorentz, and Klein models, we use the Poincaré ball model \(B_c^n=\{x \in \mathbb{R}^n \mid \|x\|^2 < r^2\}\) for its geometric property preservation. Here, \(c\) is the curvature, while \(r = \frac{1}{\sqrt{c}}\) defines the radius. The variable \(x\) denotes a position within the Poincaré ball, serving as the text embedding vector in our T2M generation model.

The Poincar\'e ball has Riemannian metric $g_x^H=(\lambda_x^c)^2g^E$ conformal to the Euclidean metric $g^E$ with $\lambda_x^c=\frac{2}{1-||x||^2}$. This exponential shrinks distances towards the boundary, allowing hierarchical structures to be represented with a large branching factor. 

We employ the exponential map $\operatorname{Exp}_x: \tau_xH\to H$ and its inverse, the logarithmic map $\operatorname{Log}_x$, to project points between the hyperbolic Poincar\'e ball and Euclidean space, where $\tau_x H$ refers to the tangent space at a point $x$ in the hyperbolic space $H$. This enables us to embed syntactic trees extracted from text as hyperbolic graphs.
Hyperbolic space is well-suited for such hierarchical data due to its ability to model low-dimensional structures with minimal distortion compared to Euclidean counterparts \citep{nickel2017poincare,leng2023dynamic}. For linguistic trees in our task, it thus provides a more suitable geometric domain.

Hyperbolic Graph Convolution (HGC) provides a powerful way to learn representations of hierarchical graph-structured data like syntactic trees in our task.
HGC generalizes graph convolutional operations to hyperbolic space by projecting node features from the ambient Euclidean space to the Poincar\'{e} ball using the exponential map Exp($\cdot$)
\citep{liu2019hyperbolic}. It then applies the $\mathcal{M}\ddot{\mathrm{o}}$bius layer operations $\otimes$ and $\oplus$ to transform features while preserving distances between embedded points \citep{kochurov2020geoopt}. 
Neighborhood aggregation is performed via hyperbolic pooling functions $\mathbf{F}^H$ to combine neighboring node representations \citep{liu2019hyperbolic}. Finally, the hyperbolic activation $\sigma^H$ introduces non-linearities by alternating between Riemannian and Euclidean spaces.
Compared to Euclidean GNNs, HGC's ability to scale vectors proportional to their hyperbolic distance allows better embedding of trees with minimal distortion. This makes it critical for modeling syntactic dependencies in our text using hyperbolic graph embeddings $\mathbb{G}=(\nu,\xi)$, where $\nu$ denotes all the nodes, specifically referring to each word in a sentence, and $\xi$ represents all the edges, specifically indicating the syntactic relationships between each pair of words. HGC thus provides a powerful way to learn task-specific representations of our hierarchical input data.

% The Hyperbolic Graph Convolution (HGC) process is defined as mapping node features $\mathbf{x}_{i}^{E}$ from the ambient Euclidean space to the hyperbolic Poincaré ball via the exponential map $\mathbf{x}_{i}^{H}= \mathrm{Exp}(\mathbf{x}_{i}^{E})$ (\cite{yang2022hyperbolic}). 
% %
% It then applies the $\mathcal{M}\ddot{\mathrm{o}}$bius layer $\mathbf{Y}^{H}_{i}= \mathbf{F}^{H}(\mathbf{x}_{i}^{H} \otimes \mathbf{W} \oplus \mathbf{b})$ utilizing operations like multiplication $\otimes$ and addition $\oplus$ with trainable parameters $\mathbf{W}, \mathbf{b}$. 
% %
% Finally, a hyperbolic activation $\mathbf{Y}_{i}^{H'}= \sigma^{H} \mathbf{Y}^{H}_{i}$ introduces non-linearities, producing the output features $\mathbf{Y}^{H'}_{i}$ in hyperbolic space. By generalizing graph convolutions via differentiable hyperbolic operations, HGC effectively encodes hierarchical structures. This enables Fg-T2M++ to capture the fine-grained dependencies in language and generate fine-grained human motions.
\begin{figure*}[t]
    \centering
    \includegraphics[width=0.88\linewidth]{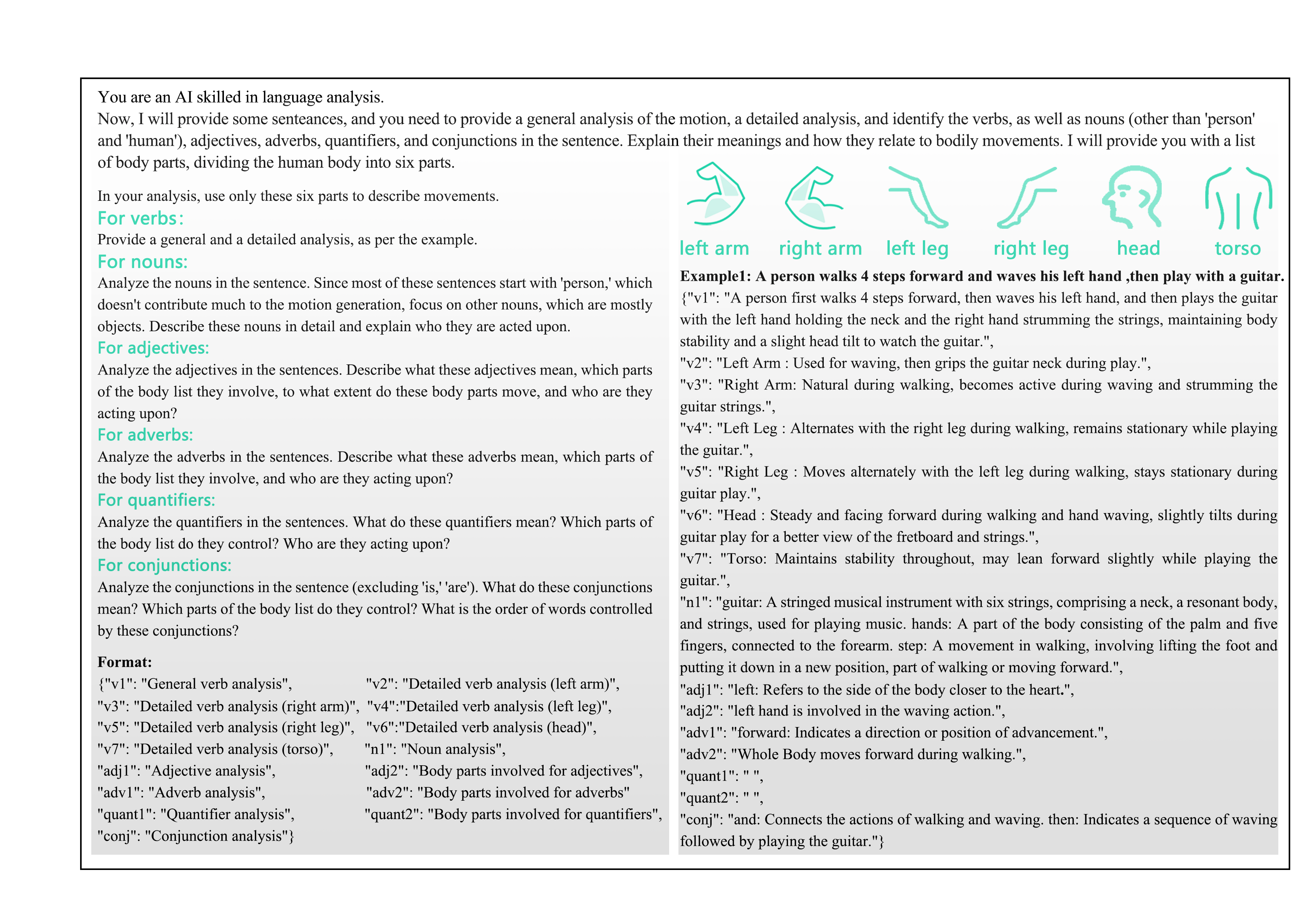}
    \caption{The prompt of strategy and example for LLMs Semantic Parsing.}
    \label{GPTprompt}
\end{figure*}

\section{Methodology}\label{sec3}

Given a text prompt, $\mathbf{W} = \{w_1,w_2,\dots,w_N\} $, $\mathbf{W}  \in \mathbb{R}^{N \times L}$ where $N$ represents the number of words and  $L$ is the dimension of word vector. Our goal is to generate a human motion sequence, denoted as $\mathbf{M} = \{m_1,m_2,\dots m_S\}$, where $\mathbf{M} \in \mathbb{R}^{S \times D}$. Here, $S$ refers to the motion sequence length and $D$ is the pose representation dimension. To achieve this, we present a diffusion model-based framework, Fg-T2M++, for fine-grained and high-fidelity text-driven motion generation. In the following sections, we provide: an overview of our motion generation approach in Section~\ref{overview}; introduction to the LLMs Semantic Parsing in Section~\ref{LASP}; the Hyperbolic Text Representation Module in Section~\ref{HLSAM}; and the Multi-Modal Fusion Module in Section~\ref{RPPR}.

\subsection{Motion Generation via Diffusion Models}\label{overview}

Our Fg-T2M++ approach generates fine-grained motions conditioned on natural language using a diffusion probabilistic model. As shown in Figure \ref{pipeline}, we sample random noise and input it to the diffusion model along with timestep $T$ and text condition $c$. The model iterates backwards from $X_T$ to $X_0$, removing noise at each step. 
Crucially, at each denoising step the text is first parsed by LLMs into fine-grained part-level descriptions. The HTP module then encodes the text using a hyperbolic linguistics tree representation. Finally, the MMF module collaboratively reasons over the noisy motion and rich text encodings to acquire the clean motion embedding.

We employ classifier-free diffusion guidance \citep{ho2022classifier} to scale conditional and unconditional distributions as:
\begin{equation}  
\epsilon = s\epsilon_\theta(x_t,t,c)+(1-s)\epsilon_\theta(x_t,t,\varnothing)
\end{equation}

\noindent where guidance scale $s$ controls the text conditioning. 
Our objective predicts the clean state $X_0$ by minimizing the L2 loss between predicted and ground truth motions, enabling Fg-T2M++ to learn high-quality generation via:
\begin{equation}
\mathcal{L}=\mathrm{E} [\parallel \mathbf{x}_0 - \epsilon_\theta(\mathbf{x}_t,t,c) \parallel_2^2]  
\end{equation}
where $\epsilon_\theta(\mathbf{x}_t,t,c)$ denotes the model predict output. This diffusion formulation empowers our approach for the challenging task.

\subsection{LLMs Semantic Parsing}\label{LASP}

Existing datasets provide rich motion data but have coarse text prompts limited to brief action descriptions like ``a person is jumping" \citep{plappert2016kit,guo2022generating}, which may overlook fine-grained information of other body parts, such as hand waving details. The absence of such detailed descriptions significantly hinders fine-grained motion generation.
Prior work encodes only high-level semantics from coarse prompts using shallow representations. As a result, generated motions do not precisely match detailed language specifications, such as coordinated part-level movements over time.
We aim to synthesize motion from fine-grained natural language descriptors. However, current methods cannot comprehend such rich descriptions.

Large language models have advanced NLP through powerful modeling, e.g. OpenAI's GPT \citep{achiam2023gpt}. Our LLMs Semantic Parsing approach leverages this by parsing prompts into annotations of part motions and semantics (e.g. nouns, adjectives, adverbs, etc.).  
This fine-grained parsing captures specifications enabling intricate linguistic-kinematic modeling. It closes the gap between coarse data and our goal of animating complex language descriptions through state-of-the-art techniques.

Our approach utilizes strong priors from LLMs like GPT-3.5 to precisely capture relationships between natural language and human motion at a fine-grained level. 
For the given text, we perform dual parsing of action and semantics. To parse action, we represent the human skeleton from SMPL \citep{loper2023smpl} and MMM \citep{terlemez2014master} as six main body parts - left arm, right arm, left leg, right leg, head, and torso. 
Leveraging GPT's understanding of language and motion knowledge, we split whole-body movements described in the text into sub-joint motions of individual parts. Inspired by ActionGPT \citep{kalakonda2023action} and GraphMotion \citep{jin2024act}, verbs are further clarified due to their decisive role in sentences.

We also parse semantics by focusing on parts of speech like nouns, adjectives, adverbs, quantifiers, and conjunctions, which convey richer meanings than other words. For instance, we interpret how adjectives and adverbs modify specific joints and how conjunctions connect actions. 
In total, each prompt undergoes 15 sub-analyses providing comprehensive parsing of both action and semantics. As Figure \ref{GPTprompt} illustrates, this dual fine-grained strategy allows comprehending motion sequence details at a granular level from the text.

\begin{figure}[t]
    \centering
    \includegraphics[width=\linewidth]{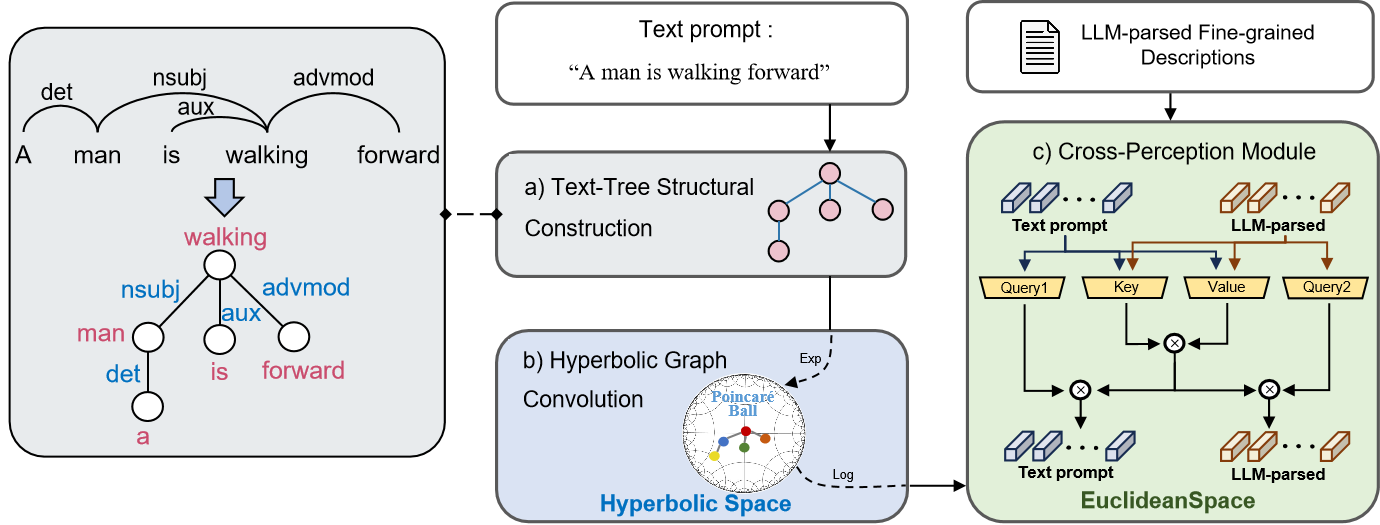}
    \caption{\textbf{Architecture of HTP}. a): the process of text-tree structural construction by dependency analysis. b): the process of hyperbolic graph convolution in the hyperbolic space to grasp the texts' precise features. c): the process of cross-perception module to make full use of the LLMs-parsed fine-grained descriptions.}
    \label{HLSA}
\end{figure}
\subsection{Hyperbolic Text Representation}\label{HLSAM}

Prior work in language-guided motion generation encodes text directly using Transformers \citep{vaswani2017attention,zhang2022motiondiffuse,tevet2023human,chen2023executing}. While capturing high-level semantics, this approach struggles to model the fine-grained details needed to comprehend language fully. 
However, text inherently contains a rich hierarchical structure, with word interconnections providing contextual information beyond standalone semantics. This structural information is not sufficiently utilized by existing methods.
To address this, we propose the Hyperbolic Text Representation Module (HTP) to leverage the inherent syntactic structure of language prompts. HTP extracts and utilizes this structural knowledge via components like text-tree construction and hyperbolic graph networks. By incorporating the additional contextual cues provided by linguistic structure, HTP facilitates improved comprehension over prior semantic encoding-focused approaches. It aims to generate motion more precisely synchronized to prompt specifications.

\paragraph{Text-Tree Structural Construction.} 
We leverage dependency parsing to identify syntactic relationships between phrases and address issues in prior work. Dependency parsing analyzes parts of speech and dependencies between words using spaCy, an NLP library for text processing functions like syntactic parsing \citep{honnibal2017spacy}. 
For a given prompt, dependency parsing establishes a text relationship tree where each word is a node and dependencies form connecting edges (Figure \ref{HLSA}a). This tree serves as the initialization for a graph, providing prior knowledge of its topological structure compared to methods lacking syntactic context modeling.
The tree represents phrases and their interdependencies more comprehensively than isolated words. This fine-grained structural information facilitates a deeper understanding of the text beyond singular semantics. By constructing linguistic trees from prompts, we aim to generate motion sequences synchronized more precisely to language descriptions.

\paragraph{Hyperbolic Graph Convolution.} 
HGC is well-suited for processing tree-structured linguistic data in hyperbolic versus Euclidean space, as hyperbolic geometry preserves local structure with low distortion  \citep{yang2022hyperbolic}. We construct a graph from the dependency parsed text tree, where words are nodes and dependencies are directed edges. Text embeddings $\mathbf{W}^E$ extracted from CLIP \citep{radford2021learning} and edge relationships $\xi$ are combined to construct the graph $\mathbf{G}^{E}=\{ \mathbf{W}^E, \xi \}$, where $E$ represents Euclidean space. This graph is projected into the Poincar\'e ball hyperbolic model via the exponential function Exp:
$\mathbf{G}^H=\operatorname{Exp}(\mathbf{G}^E)$,
where $H$ represents hyperbolic space.

Within this hyperbolic manifold, stacked HGC layers process the graph through $\mathcal{M}\ddot{\mathrm{o}}$bius calculus and hyperbolic nonlinear activations $\sigma$\textsuperscript{H}. This updates node features to capture hierarchical structure (Figure \ref{HLSA}b).
Features are then projected back to Euclidean space using the inverse exponential function Log:
\begin{equation}
\mathbf{W}^{h}= \operatorname{Log} (\sigma^{H}( \operatorname{\mathcal{M}\ddot obius} (\operatorname{Exp}(\mathbf{W}^{E}))))
\end{equation}

By applying HGC to model the linguistic structure, we obtain text encodings informed by both dependencies and syntax contexts.

\paragraph{Cross-Perception Module.} 
In the parsing phase, our LLMs-Augmented approach precisely captures movement and linguistic details from prompts, allowing deep understanding. For parsed sentences, we use CLIP to obtain initial encodings and enhance them by passing through a Transformer layer, obtaining $\mathbf{W}^{l}$. For full text prompts, our HGC captures structural semantics at multiple levels, outputting encodings. These are also enhanced by a Transformer layer, resulting in $\mathbf{W}^{t}$.

The Cross-Perception Module aims to enhance text representations by modeling relationships between the parsed and text prompts encodings. As shown in Figure \ref{HLSA}c, it applies a multi-stage attention process. Inspired by efficient attention \citep{Shen_2021_WACV}, global context features $\mathbf{F}$ are computed using key-value attention over the concatenated encodings:
\begin{equation}
\mathbf{F} = \operatorname{softmax}(\mathbf{Key} [\mathbf{W}^{l};\mathbf{W}^{t}]) \otimes (\mathbf{Value} [\mathbf{W}^{l};\mathbf{W}^{t}]),
\end{equation}
where $[\cdot ; \cdot]$ indicates a concatenation of two tensors.
Cross-attention is then applied using query vectors $\mathbf{Q}^{l}$ and $\mathbf{Q}^{t}$:
\begin{equation}
  \begin{aligned}
   &\mathbf{W}^{l}=\mathbf{W}^{l} + \operatorname{softmax}(\mathbf{Q}^{l} \mathbf{W}^{l}) \otimes \mathbf{F}, \\
   &\mathbf{W}^{t}=\mathbf{W}^{t} + \operatorname{softmax}(\mathbf{Q}^{t} \mathbf{W}^{t}) \otimes \mathbf{F}.
  \end{aligned}
\end{equation}

The enriched encodings capture multi-level relationships to guide motion generation.

\subsection{Multi-Modal Fusion}\label{RPPR}
Existing methods that learn fixed word features struggle to capture high-order semantics \citep{tevet2023human,zhang2022motiondiffuse,zhang2023remodiffuse}. 
However, human sentence comprehension proceeds hierarchically from coarse to fine. To better model this, we introduce a coarse-to-fine structure in our motion diffusion model comprising two semantic levels: overall and detailed information.

\begin{figure}[t]
    \centering
    \includegraphics[width=\linewidth]{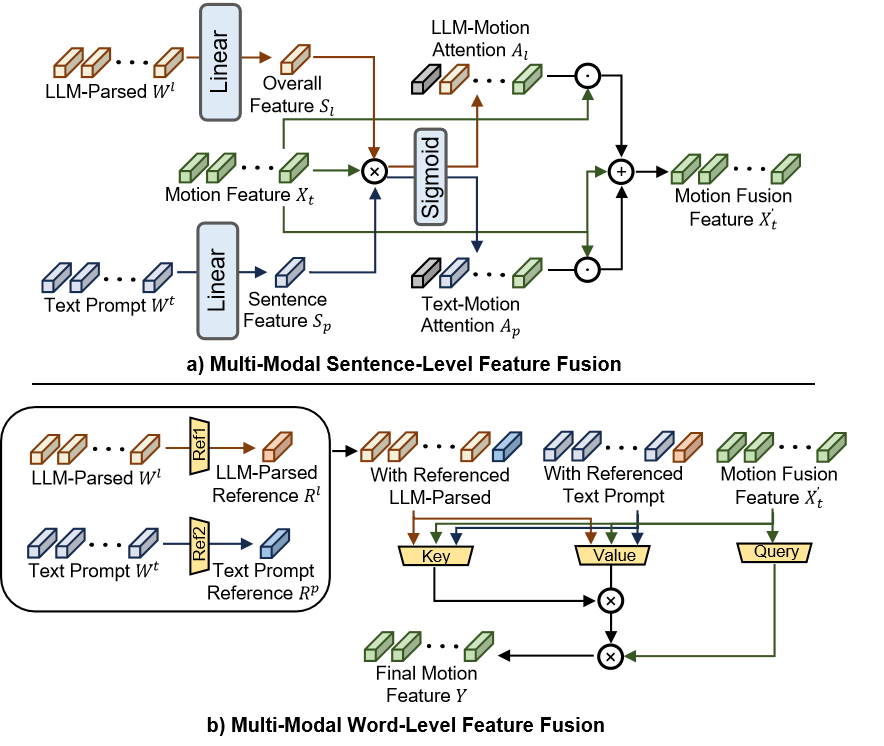}
    \caption{\textbf{Illustration of two fusion methods in MMF.} a) multi-modal sentence-level feature fusion and b) multi-modal word-level feature fusion.}
    \label{PRRP}
\end{figure}

Our proposed Multi-Modal Fusion (MMF) module aims to iteratively refine the interaction between text and motion encodings for controlling fine-grained motion diffusion. As shown in Figure \ref{PRRP}, it contains two parts:

\begin{enumerate}
\item Multi-modal sentence-level feature fusion captures the overall semantic meaning across encoded text and motion modalities.

\item Multi-modal word-level feature fusion iteratively refines text-motion features through attention to a reference sequence, given the overall semantic context computed above.
\end{enumerate}

Within our hierarchical Fg-T2M++ generation framework, MMF utilizes encoded representations from the Cross-Perception module. It provides progressively refined text signals to guide motion diffusion from coarse to fine-grained details. This hierarchical reasoning approach allows better modeling of language at both global and local levels, addressing challenges in prior work with fixed encodings.

\paragraph{Multi-Modal Sentence-Level Feature Fusion} 

aims to combine multi-modal control signals from parsed content and text prompts at the overall semantic level. As illustrated in Figure \ref{PRRP}a, we first transform the parsed content into overall features $S_l$ and the text prompts into sentence features $S_p$.
We then calculate frame-level attention maps $A_l$ and $A_p$ denoting relevance between each motion feature $X_t$ and the sentence features:
\begin{equation}
\mathbf{A}_{l}= \mathbf{X}_{t} (\mathbf{S}^{l})^T, \mathbf{A}_{p}= \mathbf{X}_{t} (\mathbf{S}^{p})^T
\end{equation}

This captures correspondence between visual motion and linguistic semantics.
The cross-modal motion features $X_t'$ are obtained by highlighting channels in $X_t$ related to both sentences:
\begin{equation}
{\mathbf{X}_t^{\prime} = \mathbf{X}_t + \lambda_{l} (\mathbf{X}_t \odot \sigma(\mathbf{A}_{l})) + \lambda_{p} (\mathbf{X}_t \odot \sigma(\mathbf{A}_{p}))}
\end{equation}

\noindent where $\lambda_{l}$ and $\lambda_{p}$ terms control contribution and $\sigma$ is the sigmoid activation. This fusion operates at the coarse semantic level to provide context for the following word-level feature refinement.

\paragraph{Multi-Modal Word-Level Feature Fusion} 

captures fine-grained text-motion relationships. Inspired by ReMoDiffuse~\citep{zhang2023remodiffuse}, it iteratively refines encodings through hybrid attention to a shared reference sequence. Specifically, at each timestep the encodings act as $\mathbf{Queries}$ while the reference acts as $\mathbf{Keys}$ and $\mathbf{Values}$. We employ a hybrid attention combining self-attention and cross-attention~\citep{vaswani2017attention} to compute the interaction. This extracts dependencies between words by relating them to context. Crucially, it operates on complementary LLMs-parsed and text prompt encodings, enabling informative exchange. Through iterative hybrid attention refinement, the module provides control signals capturing detailed semantic associations to guide high-fidelity generation of motion sequences conditioned on natural language description. 

This fusion component aims to capture cross-modal relationships via iterative refinement of text-motion representations guided by contextual relationships extracted using hybrid attention computations. We first obtain sentence-level features $\mathbf{S}^l$ and $\mathbf{S}^t$ encoding overall semantics of the LLMs and text prompt inputs, respectively. Reference representations $\mathbf{R}^l$ and $\mathbf{R}^t$ are then generated from $\mathbf{S}^l$ and $\mathbf{S}^t$ using trainable projection matrices $\mathbf{M^l}$ and $\mathbf{M^t}$:
\begin{equation}
\mathbf{R}^{l}=\mathbf{M}^{l} \mathbf{S}^{l}, \;\;\;   \mathbf{R}^{t}=\mathbf{M}^{t} \mathbf{S}^{t}
\end{equation}

We concatenate word features $\mathbf{W}^l$, $\mathbf{W}^t$ with their respective references $\mathbf{R}^t$, $\mathbf{R}^l$ and motion features $\mathbf{X}_t^{\prime}$ to compute unified $\mathbf{Keys}$ and $\mathbf{Values}$ for hybrid attention. This models their mutual influence through cross-attending references:
\begin{equation}
\begin{aligned}
&\mathbf{Value} = [\mathbf{V}^{m} \mathbf{X}_t^{\prime}; \mathbf{V}^{l} [\mathbf{W}^l; \mathbf{R}^{t}]; \mathbf{V}^{t} [\mathbf{W}^t; \mathbf{R}^{l}]],\\
&\mathbf{Key} = [\mathbf{K}^{m} \mathbf{X}_t^{\prime}; \mathbf{K}^{l} [\mathbf{W}^l; \mathbf{R}^{t}]; \mathbf{K}^{t} [\mathbf{W}^t; \mathbf{R}^{l}]]  
\end{aligned}
\end{equation}
where $\mathbf{V}^{m},\mathbf{V}^{l},\mathbf{V}^{t},\mathbf{K}^{m},\mathbf{K}^{l},\mathbf{K}^{t}$ denote trainable matrices.
Global templates $\mathbf{G}$ extracted via softmax attention enable iterative refinement of text-motion representations:
\begin{equation}
        \mathbf{G} = \operatorname{softmax} (\mathbf{Key})\mathbf{Value}
\end{equation}

Additionally, we generate a query vector at each refinement iteration to learn contextual relationships from the global template. Specifically, the motion encoding ${\mathbf{X}_t'}$ serves as input to produce the query vector via a trainable projection matrix ${\mathbf{Q}^m}$:
\begin{equation} 
\mathbf{Query}=\mathbf{Q}^{m} \mathbf{X}_t^{\prime}
\end{equation}

This query vector attends to the global template ${\mathbf{G}}$, which encapsulates dependencies across text and motion representations inferred through iterative computations of hybrid attention over inputs from the two modalities.
\begin{equation}
\mathbf{Y} = \operatorname{softmax} (\mathbf{Query})\mathbf{G}  
\end{equation}
where ${\mathbf{Y}\in\mathbb{R}^{S\times D}}$ gives the updated output.

By repeatedly refining encodings through extracting contextual relationships from $\mathbf{G}$, our model incrementally fuses hierarchical semantics between modalities. This enables better comprehension of textual content by grounding precise or subtle motion details in word-level semantics, enhancing performance on tasks involving understanding text through reference to implied motion concepts.

\begin{table*}[t]
\centering

\resizebox{2.105\columnwidth}{!}{%
{
\begin{tabular}{lcccccccc}
\hline

\multirow{2}{2cm}{\centering Methods} &\multirow{2}{2cm}{\centering Publication}& \multicolumn{3}{c}{\centering R Precision$\uparrow$} & \multirow{2}{1.5cm}{\centering FID$\downarrow$} & \multirow{2}{2.5cm}{\centering MultiModal Dist$\downarrow$} & \multirow{2}{2cm}{\centering Diversity$\uparrow$} & \multirow{2}{2cm}{\centering MultiModality$\uparrow$} \\
&& Top 1 & Top 2 & Top 3 \\

\hline

TEMOS \citep{petrovich2022temos}&
ECCV&
  $0.424^{\pm.002}$ &
  $0.612^{\pm.002}$ &
  $0.722^{\pm.002}$ &
  $3.734^{\pm.028}$ &
  $3.703^{\pm.008}$ &
  $8.973^{\pm.071}$ &
  $0.368^{\pm.018}$ \\
  
MDM \citep{tevet2023human}&
ICLR&
  $0.320^{\pm.005}$ &
  $0.498^{\pm.004}$ &
  $0.611^{\pm.007}$ &
  $0.544^{\pm.044}$ &
  $5.566^{\pm.027}$ &
  $9.559^{\pm.086}$ &
  $\mathbf{\textcolor{red}{2.799^{\pm.072}}}$ \\
  
  MotionDiffuse \citep{zhang2022motiondiffuse}&
  TPAMI&
    $0.491^{\pm.001}$ &
  $0.681^{\pm.001}$ &
  $0.782^{\pm.001}$ &
  $0.630^{\pm.001}$ &
  $3.113^{\pm.001}$ &
  $9.410^{\pm.049}$ &
  $1.553^{\pm.042}$ \\
  
  Temporal VAE \citep{guo2022generating}&
  CVPR&
    $0.455^{\pm.003}$ &
  $0.636^{\pm.003}$ &
  $0.740^{\pm.003}$ &
  $1.067^{\pm.002}$ &
  $3.340^{\pm.008}$ &
  $9.188^{\pm.002}$ &
  $2.090^{\pm.083}$ \\
  
MLD \citep{chen2023executing}&
CVPR&
  $0.481^{\pm.003}$ &
  $0.673^{\pm.003}$ &
  $0.772^{\pm.002}$ &
  $0.473^{\pm.013}$ &
  $3.196^{\pm.010}$ &
  ${9.724}^{\pm.082}$ &
  $2.413^{\pm.079}$ \\

   T2M-GPT \citep{zhang2023generating}&
   CVPR&
    $0.491^{\pm.003}$ &
  $0.680^{\pm.003}$ &
   $0.775^{\pm.002}$ &
    $0.116^{\pm.004}$ &
     $3.118^{\pm.011}$ &
      $\mathbf{\textcolor{red}{9.761^{\pm.081}}}$ &
       $1.856^{\pm.011}$ \\
  MotionGPT \citep{jiang2024motiongpt}&
    NeurIPS&
  ${0.492}^{\pm.003}$ &
  $0.681^{\pm.003}$ &
  $0.778^{\pm.002}$ &
  $0.232^{\pm.008}$ &
  ${3.096}^{\pm.008}$ &
  $9.528^{\pm.071}$ &
  $2.008^{\pm.084}$ \\
GraphMotion \citep{jin2024act}&
NeurIPS&
$0.504^{\pm.003}$ &
$0.699^{\pm.002}$ &
$0.785^{\pm.002}$ &
$0.116^{\pm.007}$ &
$3.070^{\pm.008}$ &
$9.692^{\pm.067}$ &
$2.766^{\pm.096}$ \\
FineMoGen \citep{zhang2024finemogen}&
NeurIPS&
$0.504^{\pm.002}$ &
$0.690^{\pm.002}$ &
$0.784^{\pm.002}$ &
$0.151^{\pm.008}$ &
$2.998^{\pm.008}$ &
$9.263^{\pm.094}$ &
$2.696^{\pm.079}$ \\
Att-T2M \citep{zhong2023attt2m}&
ICCV&
$0.499^{\pm.003}$ &
$0.690^{\pm.002}$ &
$0.786^{\pm.002}$ &
$0.112^{\pm.006}$ &
$3.038^{\pm.007}$ &
$9.700^{\pm.090}$ &
$2.452^{\pm.051}$ \\
        ReMoDiffuse \citep{zhang2023remodiffuse}&
ICCV&
$0.510^{\pm.005}$ &
$0.698^{\pm.006}$ &
$0.795^{\pm.004}$ &
$0.103^{\pm.004}$ &
$2.974^{\pm.016}$ &
$9.018^{\pm.075}$ &
$1.795^{\pm.043}$ \\
Fg-T2M \citep{wang2023fg}&
ICCV&
$0.492^{\pm.002}$ &
$0.683^{\pm.003}$ &
$0.783^{\pm.002}$ &
$0.243^{\pm.019}$ &
$3.109^{\pm.007}$ &
$9.278^{\pm.072}$ &
$1.614^{\pm.049}$ \\

\hline

\cellcolor{yellow!15}Fg-T2M++ 
&\cellcolor{yellow!15} - 
&\cellcolor{yellow!15}$\mathbf{\textcolor{red}{0.513^{\pm.002}}}$ 
&\cellcolor{yellow!15}$\mathbf{\textcolor{red}{0.702^{\pm.002}}}$ 
& \cellcolor{yellow!15}$\mathbf{\textcolor{red}{0.801^{\pm.003}}}$ 
& \cellcolor{yellow!15}$\mathbf{\textcolor{red}{0.089^{\pm.004}}}$ 
& \cellcolor{yellow!15}$\mathbf{\textcolor{red}{2.925^{\pm.007}}}$ 
& \cellcolor{yellow!15}$9.223^{\pm.114}$ 
& \cellcolor{yellow!15}$2.625^{\pm.084}$\\

\hline

\end{tabular}}}

\caption{\textbf{Quantitative evaluation on the HumanML3D \citep{guo2022generating} test set.} We run all the evaluation 20 times and $\pm$ indicates the 95\% confidence interval. \textcolor{red}{Red} indicates the best result.}
\label{humanml3d}
\end{table*}

\begin{table*}[t]
\centering

\resizebox{2.105\columnwidth}{!}{%
{
\begin{tabular}{lcccccccc}
\hline

\multirow{2}{2cm}{\centering Methods} &\multirow{2}{2cm}{\centering Publication}& \multicolumn{3}{c}{\centering R Precision$\uparrow$} & \multirow{2}{1.5cm}{\centering FID$\downarrow$} & \multirow{2}{2.5cm}{\centering MultiModal Dist$\downarrow$} & \multirow{2}{2cm}{\centering Diversity$\uparrow$} & \multirow{2}{2cm}{\centering MultiModality$\uparrow$} \\
&& Top 1 & Top 2 & Top 3 \\

\hline

TEMOS \citep{petrovich2022temos}&
    ECCV &
  $0.353^{\pm.006}$ &
  $0.561^{\pm.007}$ &
    $0.687^{\pm.005}$ & 
    $3.717^{\pm.051}$ & 
    $3.417^{\pm.019}$ & 
    $10.84^{\pm.100}$ & 
    $0.532^{\pm.034}$ \\
  MDM \citep{tevet2023human}&
  ICLR &
    $0.164^{\pm.004}$ &
  $0.291^{\pm.004}$ &
  $0.396^{\pm.004}$ &
  $0.497^{\pm.021}$ &
  $9.190^{\pm.022}$ &
  $10.85^{\pm.109}$ &
  $1.907^{\pm.214}$ \\
  MotionDiffuse \citep{zhang2022motiondiffuse}&
TPAMI&
  $0.417^{\pm.004}$ &
  $0.621^{\pm.004}$ &
  $0.739^{\pm.004}$ &
  $1.954^{\pm.062}$ &
  $2.958^{\pm.005}$ &
  $11.10^{\pm.143}$ &
  $0.730^{\pm.013}$ \\
  Temporal VAE \citep{guo2022generating}&
CVPR&
  $0.361^{\pm.006}$ &
  $0.559^{\pm.007}$ &
  $0.693^{\pm.007}$ &
  $2.770^{\pm.109}$ &
  $3.401^{\pm.008}$ &
  $10.91^{\pm.119}$ &
  $1.482^{\pm.065}$ \\
  MLD \citep{chen2023executing}&
  CVPR &
    $0.390^{\pm.008}$ &
  $0.609^{\pm.008}$ &
  $0.734^{\pm.007}$ &
  $0.404^{\pm.027}$ &
  $3.204^{\pm.027}$ &
  $10.80^{\pm.117}$ &
 $2.192^{\pm.071}$ \\
        T2M-GPT \citep{zhang2023generating}&
        CVPR &
    $0.416^{\pm.006}$ &
  $0.627^{\pm.006}$ &
   $0.745^{\pm.006}$ &
    $0.514^{\pm.029}$ &
     $3.007^{\pm.023}$ &
      $10.92^{\pm.108}$ &
       $1.570^{\pm.039}$ \\
  MotionGPT \citep{jiang2024motiongpt}&
  NeurIPS &
  $0.366^{\pm.005}$ &
  $0.558^{\pm.004}$ &
  $0.680^{\pm.005}$ &
  $0.510^{\pm.016}$ &
  $3.527^{\pm.021}$ &
  $10.35^{\pm.084}$ &
  $2.328^{\pm.117}$ \\
GraphMotion \citep{jin2024act}&
NeurIPS&
$0.429^{\pm.007}$ &
$0.648^{\pm.006}$ &
$0.769^{\pm.006}$ &
$0.313^{\pm.013}$ &
$3.076^{\pm.022}$ &
$\mathbf{\textcolor{red}{11.12^{\pm.135}}}$ &
$\mathbf{\textcolor{red}{3.627^{\pm.113}}}$ \\
FineMoGen \citep{zhang2024finemogen}&
NeurIPS&
$0.432^{\pm.006}$ &
$0.649^{\pm.005}$ &
$0.772^{\pm.006}$ &
$0.178^{\pm.007}$ &
$2.869^{\pm.014}$ &
$10.85^{\pm.115}$ &
$1.877^{\pm.093}$ \\
Att-T2M \citep{zhong2023attt2m}&
ICCV&
$0.413^{\pm.006}$ &
$0.632^{\pm.006}$ &
$0.751^{\pm.006}$ &
$0.870^{\pm.039}$ &
$3.039^{\pm.021}$ &
$10.96^{\pm.123}$ &
$2.281^{\pm.047}$ \\
            ReMoDiffuse \citep{zhang2023remodiffuse}&
            ICCV &
   $0.427^{\pm.014}$ &
   $0.641^{\pm.004}$ &
   $0.765^{\pm.055}$ &
     $0.155^{\pm.006}$ &
      $2.814^{\pm.012}$ &
      $10.80^{\pm.105}$ &
       $1.239^{\pm.028}$ \\

Fg-T2M \citep{wang2023fg}&
ICCV&
$0.418^{\pm.005}$ &
$0.626^{\pm.004}$ &
$0.745^{\pm.004}$ &
$0.571^{\pm.047}$ &
$3.114^{\pm.015}$&
$10.93^{\pm.083}$ &
$1.019^{\pm.029}$\\

\hline

\cellcolor{yellow!15}Fg-T2M++
&\cellcolor{yellow!15} - 
&\cellcolor{yellow!15}$\mathbf{\textcolor{red}{0.442^{\pm.006}}}$ 
&\cellcolor{yellow!15}$\mathbf{\textcolor{red}{0.657^{\pm.005}}}$ 
& \cellcolor{yellow!15}$\mathbf{\textcolor{red}{0.781^{\pm.004}}}$ 
& \cellcolor{yellow!15}$\mathbf{\textcolor{red}{0.135^{\pm.004}}}$ 
& \cellcolor{yellow!15}$\mathbf{\textcolor{red}{2.696^{\pm.011}}}$ 
& \cellcolor{yellow!15}$10.99^{\pm.105}$ 
& \cellcolor{yellow!15}$1.255^{\pm.078}$\\
\hline
\end{tabular}}}
\caption{\textbf{Quantitative evaluation on the KIT-ML \citep{plappert2016kit} test set.}}
\label{kit}
\end{table*}

\section{Experiments}\label{sec4}

\subsection{Experimental Settings}

\paragraph{Dataset.} There exist some datasets for conditional generation, such as those proposed in \citep{plappert2016kit,guo2022generating,guo2020action2motion,punnakkal2021babel}. However, datasets like \citep{punnakkal2021babel} and \citep{guo2020action2motion}, based on action categories, cannot provide complete textual sentences, making it unsuitable for analyzing the intrinsic connections of sentence structure for our method. Instead, we use the datasets, particularly the HumanML3D dataset \citep{guo2022generating} and the KIT-ML Motion-Language dataset \citep{plappert2016kit} for experiments. The HumanML3D dataset \citep{guo2022generating} is a combination of HumanAct12 \citep{guo2020action2motion} and AMASS \citep{mahmood2019amass} datasets, each motion described by 3 text scripts, with an average length of about 12 words. The HumanML3D dataset \citep{guo2022generating} includes 14616 motions and 44970 text descriptions, involving various human activities such as daily activities, sports, acrobatics, etc., with a total duration of approximately 28.59 hours. The KIT Motion-Language dataset \citep{plappert2016kit} provides a smaller-scale evaluation benchmark, with each motion sequence accompanied by one to four sentences, averaging 8 words in description length. The KIT-ML dataset \citep{plappert2016kit} consists of 3911 motion sequences and 6353 natural language descriptions, totaling approximately 10.33 hours.

\paragraph{Evaluation Metrics.} Following the evaluation metrics \citep{guo2022generating}. (1) R-Precision (R-TOP). For each inferred text-motion pair, 31 unmatched descriptions are randomly selected from the test set. The average top-k precision is obtained by calculating and ranking the Euclidean distance between the motion and each of the 32 descriptions. (2) Frechet Inception Distance (FID). FID measures the similarity between the feature distributions extracted from the generated motions and the ground truth motions. (3) MultiModal Distance (MM-Dist). For a given description, the multimodal distance between the textual features and the corresponding generated motion features is calculated. (4) Diversity. Diversity evaluates the dissimilarity among all generated motions across all descriptions by calculating the average pairwise Euclidean distance between randomly partitioned groups of motions. (5) Multimodality. For a given text description, 32 motion sequences are randomly generated, and multimodality quantifies the dissimilarity among these generated motion sequences. We primarily focus on R-Precision and FID as key performance indicators, as they are important metrics for assessing the overall quality of generated motions.

\paragraph{Implementation Details.} 
Regarding the motion encoder, we employ a 4-layer transformer with a latent dimension of 512. As for the text encoder, a frozen text encoder from CLIP ViT-B/32 is utilized, complemented by two additional transformer encoder layers. In terms of the diffusion model, the variances $\beta_t$ are predefined to linearly spread from 0.0001 to 0.02, and the total number of noising steps is set at T = 1000. We use the Adam optimizer to train the model with an initial learning rate of 0.0002, gradually decreasing to 0.00002 through a cosine learning rate scheduler. The training process is conducted on 4 NVIDIA GeForce RTX 3090, with a batch size of 256 on a single GPU.

\begin{figure*}[t]
    \centering
    \includegraphics[width=0.9\linewidth]{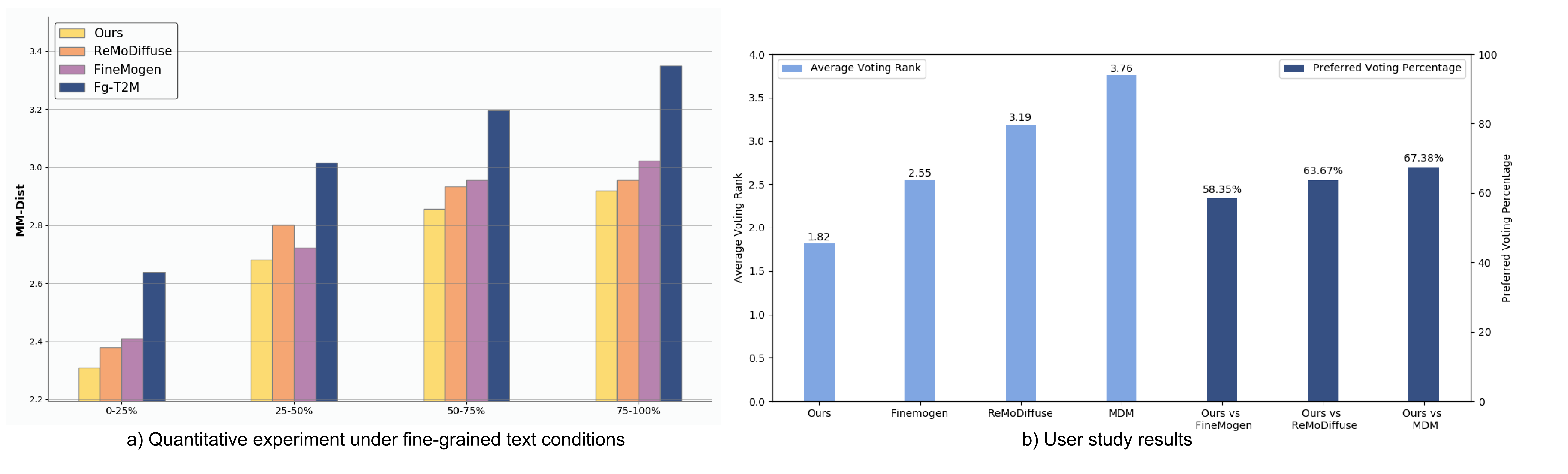}
    \caption{\textbf{Fine-Grained Evaluation Experiment Results.} a) Evaluation of MM-Dist based on different numbers of fine-grained POSs on KIT-ML datasets \citep{plappert2016kit}, where lower MM-Dist indicates better performance. The range from 0-25\% to 75-100\% signifies increasing difficulty levels. b) User study results on HumanML3D datasets \citep{guo2022generating}. The light blue bars on the left indicate the average voting rankings for each method, with lower rankings being better. The dark blue bars on the right represent the preference rate of Fg-T2M++ compared to other models, with higher values being better.}
    \label{Finegrained}
\end{figure*}
For pose representation $D$, we follow Guo \etal \citep{guo2022generating}. The pose states contain seven different parts: ($r^{va}, r^{vx}, r^{vz}, r^{h}, j^{p}, j^{v}, j^{r}$). Here $r^{va} \in \mathbb{R}$ is the root joint`s angular velocity along the Y-axis, $r^{vx}, r^{vz} \in \mathbb{R}$ are the root joint`s linear velocities along the X-axes and Z-axes, respectively. $r^h \in \mathbb{R}$ is the height of the root height. $j^p, j^v \in \mathbb{R}^{J\times3}$ are the positions and linear velocities of each joints. $j^r \in \mathbb{R}^{J\times6}$ is the 6D rotation of each joint. $J$ represents the number of joints, which are 22 in HumanML3D dataset \citep{guo2022generating} and 21 in KIT-ML dataset \citep{plappert2016kit}.

\subsection{Comparison with the State of the Art}

We compared our method with state-of-the-art (SOTA) models \citep{petrovich2022temos,guo2022generating,zhang2022motiondiffuse,tevet2023human,chen2023executing,zhang2023generating,jiang2024motiongpt,jin2024act,zhang2024finemogen,zhong2023attt2m,zhang2023remodiffuse,wang2023fg}. Quantitative comparisons of our method with these models on the HumanML3D \citep{guo2022generating} and KIT-ML \citep{plappert2016kit} datasets are shown in Table \ref{humanml3d} and \ref{kit}, respectively.

\begin{table}[t]
\centering

\begin{tabular}{>{\raggedright\arraybackslash}p{0.15\linewidth} >{\raggedright\arraybackslash}p{0.75\linewidth}}
\toprule
\textbf{Amount of POS} & \textbf{Sample Prompts} \\
\midrule
Tail  &-  A person does a jumping jack. \\
0-25\% &- The person runs forward fast. \\
\midrule
Tail  &- A man jogs and stops. \\
25-50\% &- The person kicked with left leg. \\
\midrule
Tail 50-75\%&- A man bends to his left several times while stretching his right arm over his head. \\
 &- The person runs to their left then curves to the right and continues to run then stops. \\
\midrule
Tail 75-100\%&- A person jumps and brings both arms above his head as he spread his legs and then moves them back into the original position. \\
 &- The man takes a step and picks up 3 things takes a few more steps and places one thing on the table then turns around to head back. \\
\bottomrule

\end{tabular}
\caption{Sample prompts showcasing different amount of fine-grained part-of-speech within the descriptions.}
\label{fine_descriptions}
\end{table}

Compared to other methods, Fg-T2M++ achieves significantly higher scores in R-TOP, FID, and MM-Dist. These results highlight our method's proficiency in generating high-quality motion sequences that seamlessly align with the intended meanings of the provided textual prompts. Compared to SOTAs, our approach demonstrates superior performance across accuracy metrics, including R-TOP, FID, and MM-Dist. Notably, when compared to ReMoDiffuse \citep{zhang2023remodiffuse}, which employs a motion retrieval-augmented generation method aimed at matching ground truth motion distributions, our Fg-T2M++ stands out. This is attributed to our innovative approach, leveraging a sentence analysis module and LLMs parsing, which enables generated motions to better align with textual prompts. Remarkably, even without additional ground truth motion priors, Fg-T2M++ consistently outperforms competitors across all precision metrics.

In terms of diversity metrics such as MultiModality and Diversity, the fine-grained guidance provided by our LLMs-parsed model tends to prioritize strict adherence to textual semantics. While this results in slightly weaker performance on diversity metrics, it ensures that generated motions align closely with expected textual prompts. It is important to note that prioritizing accuracy metrics strengthens the persuasiveness of our approach. After all, if generated motions fail to align with expected results, diversity metrics lose their significance. Overall, our method demonstrates advanced experimental results and showcases the robustness of our model's performance across both datasets.

Note that when applied to general measurement methods for T2M generation, the commonly used metrics may appear moderate. This is because, when compared to real data, state-of-the-art methods achieved close scores. Hence, these metrics might not provide precise assessments, especially for more challenging complex text conditions in the generation process. In response, we specifically conducted fine-grained evaluation experiments under complex textual conditions, which we will delve into in the next section.

\subsection{Fine-Grained Evaluation Experiments}

We designed two evaluation experiments to assess our model's fine-grained adaptability. The first is the quantitative experiments under fine-grained text conditions. As for fine-grained texts, we rank the data according to fine-grained part-of-speech (POS) counts of ``adjectives," ``adverbs," ``conjunctions," and ``quantifiers" in sentences. 
We categorize all samples by ranking POS counts in ascending order, dividing the data into 0-25\%, 25-50\%, 50-75\%, and 75-100\%, from lower to higher counts.
We offer two examples for each data range to better illustrate the complexity across various fine-grained parts of speech ranges, as depicted in Table \ref{fine_descriptions}. With higher Tail percentages, textual prompts transition from simple action sentences to complex structures containing multiple parts of speech. This evolution demands that generated motions become more fine-grained and challenging. Compared to \citep{zhang2023remodiffuse,zhang2024finemogen,wang2023fg}, our method outperforms the current SOTA methods, as shown in Figure \ref{Finegrained}a. Our method still maintains a better performance even though there are abundant fine-grained words in the 50-75\% and 75-100\% splits, indicating a better ability to capture fine-grained details.

\begin{table}[t]
    \centering
\resizebox{\columnwidth}{!}{%
    \begin{tabular}{l ccc ccc}
    \toprule  
    \multirow{2}{*}{Methods}  & \multicolumn{3}{c}{R Precision $\uparrow$}& \multirow{2}{*}{FID$\downarrow$}& \multirow{2}{*}{MM-Dist$\downarrow$}&\\
         \cmidrule{2-4}
     
    ~&R-TOP1  &R-TOP2 &R-TOP3  \\
        \hline
        Fg-T2M &$ 0.418^{\pm.005}$&$ 0.626^{\pm.004}$&$ 0.745  ^{\pm.004}$&$ 0.571 ^{\pm.047}$&$ 3.114^{\pm.015}$ \\

         \hline
    \rowcolor{lightgray}
    \multicolumn{7}{c}{\textit{\textbf{Component Analysis of LLMs Semantic Parsing}}} \\
        \hline
         Only Text Prompt &$ 0.430^{\pm.005}$&$ 0.641^{\pm.003}$&$ 0.761  ^{\pm.006}$&$ 0.344  ^{\pm.021}$&$ 2.757^{\pm.028}$  \\
        \quad+ Word Semantic Parsing (n1-conj) &$ 0.439^{\pm.010}$&$ 0.649^{\pm.010}$&$ 0.773  ^{\pm.014}$&$ 0.259 ^{\pm.088}$&$ 2.735^{\pm.013}$ \\
        \quad\quad + Action Body Parsing (v1-v7) &  $0.442 ^{\pm.006}$& $ 0.657 ^{\pm.005}$& $0.781 ^{\pm.004}$& $ 0.135 ^{\pm.004}$& $ 2.696 ^{\pm.011}$ \\

            \hline
    \rowcolor{lightgray}
    \multicolumn{7}{c}{\textit{\textbf{Component Analysis of Hyperbolic Text Representation Module}}} \\
        \hline
         Standard Transformer &$ 0.410^{\pm.007}$&$ 0.611^{\pm.004}$&$0.729^{\pm.007}$&$ 0.724 ^{\pm.043}$&$ 3.234^{\pm.019}$\\
        \quad+ GCN &$ 0.428^{\pm.005}$&$ 0.641^{\pm.004}$&$0.765^{\pm.005}$&$ 0.357 ^{\pm.036}$&$ 2.867^{\pm.011}$\\
        \quad\quad+ Hyperbolic GCN &$ 0.435^{\pm.006}$&$ 0.650^{\pm.006}$&$ 0.773 ^{\pm.005}$&$  0.164 ^{\pm.024}$&$2.725^{\pm.014}$ \\
        \quad\quad\quad+ Cross-Perception Module &  $0.442 ^{\pm.006}$& $ 0.657 ^{\pm.005}$& $0.781 ^{\pm.004}$& $ 0.135 ^{\pm.004}$& $ 2.696 ^{\pm.011}$ \\

            \hline
    \rowcolor{lightgray}
    \multicolumn{7}{c}{\textit{\textbf{Component Analysis of Multi-Modal Fusion Module}}} \\
        \hline
        
        Only Word-Level Feature Fusion &  $0.421 ^{\pm.010}$& $ 0.635 ^{\pm.011}$& $0.760  ^{\pm.008}$&  $0.281 ^{\pm.037}$&  $2.801 ^{\pm.019}$ \\
        \quad+ Word-Level Reference &  $0.426 ^{\pm.008}$& $ 0.641 ^{\pm.006}$& $0.764 ^{\pm.007}$& $ 0.225 ^{\pm.026}$& $ 2.761 ^{\pm.042}$ \\
        \quad\quad+ Sentence-Level Feature Fusion &$ 0.437^{\pm.005}$&$ 0.651^{\pm.006}$&$0.775 ^{\pm.006}$&$ 0.167^{\pm.010}$&$ 2.735^{\pm.025}$ \\
        \quad\quad\quad+ Sentence-Level Reference &  $0.442 ^{\pm.006}$& $ 0.657 ^{\pm.005}$& $0.781 ^{\pm.004}$& $ 0.135 ^{\pm.004}$& $ 2.696 ^{\pm.011}$ \\

        \midrule
        \textbf{Fg-T2M++} & $\textbf{0.442}^{\pm.006}$& $\textbf{0.657}^{\pm.005} $& $\textbf{0.781}^{\pm.004}  $& $ \textbf{0.135}^{\pm.004} $& $ \textbf{2.696}^{\pm.011} $ \\
    \bottomrule
    \end{tabular}}
    \caption{\textbf{Ablation of the proposed components.} All results are reported on the KIT-ML \citep{plappert2016kit} test set.}
    \label{ablation}
\end{table}

The second part involved a user study, where we conducted comparisons with FineMoGen \citep{zhang2024finemogen}, ReMoDiffuse \citep{zhang2023remodiffuse}, and MDM \citep{tevet2023human}. We collected average voting ranks and user preferences to validate our earlier findings. This user study engaged 30 participants, who evaluated 15 motions generated by each method, aiming to gather comparative feedback on the question ``Which method performs better in fine-grained motion modeling?''. The statistical data from the user study is presented in Figure \ref{Finegrained}b. Our FG-T2M++ achieved the best voting ranking and demonstrated superior performance in preferred voting percentage compared to SOTA methods.

All of this highlights the adaptability of our FG-T2M++, showing its robustness in generating motions and indicating a stronger capability in capturing fine-grained details even in complex fine-grained modeling situations.

\subsection{Component Analysis and Discussion}

In Table \ref{ablation}, we conducted a comprehensive evaluation of the impact of various design components within Fg-T2M++, showcasing its performance in text-to-motion generation through extensive comparisons.

\paragraph{The Effectiveness of LLMs Semantic Parsing.} We analyzed 15 sub-components of LLMs parsing in Table \ref{ablation}, with the first seven focusing on action body parsing and the remaining eight on word semantic parsing. When compared to our baseline Fg-T2M method, which utilizes only text prompts, the incorporation of LLMs analysis in Fg-T2M++ led to a significant performance enhancement. Particularly noteworthy is the greater impact of action body parsing on performance compared to word semantic parsing. Action body parsing plays a pivotal role in improving the quality of text-to-motion generation.
\begin{figure}[t]
    \centering
    \includegraphics[width=0.8\linewidth]{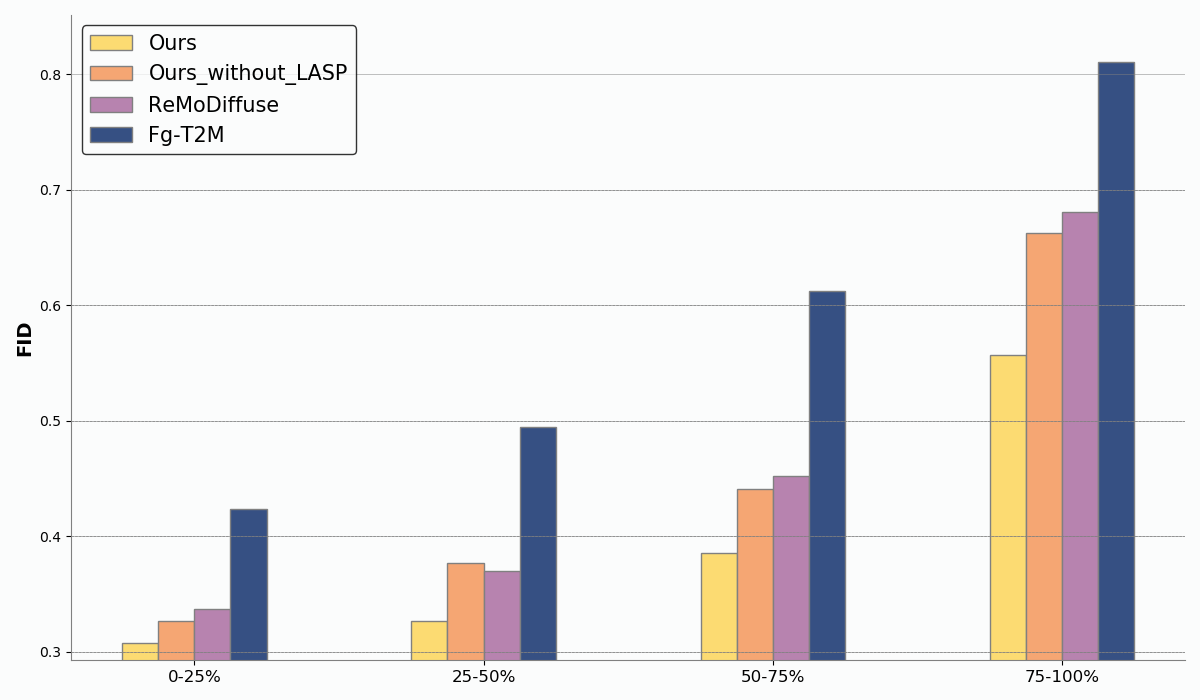}
    \caption{\textbf{Evaluation FID based on different levels of rareness on KIT-ML~\citep{plappert2016kit} datasets,} where lower FID indicates better generalization. From 0-25\% to 75-100\% signifies increasing difficulty levels.}
    \label{rare_fid}
\end{figure}

To further examine the role of LLMs Semantic Parsing, we conducted additional experiments under rare text conditions to validate the model's generalization performance. For rare texts, we followed the ReMoDiffuse \citep{zhang2023remodiffuse} metric, which introduces the concept of the sample's rareness. As for a test prompt, we calculate its rareness $r_p$ as:
\begin{equation}
    r_p = 1 - \max_i \{<E(\mathrm{text}_{i}), E(\mathrm{prompt})>\} ,
\end{equation}
where $E$ represents the CLIP text encoder, $\mathrm{text}_{i}$ is the motion description in the training set, and $<\cdot,\cdot>$ denotes cosine similarity. This formula quantifies the maximum similarity between a given prompt and motion description in the training set. The higher the similarity, the lower the rareness, and vice versa.  For rare texts, we rank the data based on their rareness value and divide the data into Tail 0-25\%, Tail 25-50\%, Tail 50-75\%, and Tail 75-100\%, ranging from common to rare. Compared with ReMoDiffuse \citep{zhang2023remodiffuse} and Fg-T2M \citep{wang2023fg}, in Figure \ref{rare_fid}, Fg-T2M++ generates motion sequences that more conform to the ground truth distribution, especially under rarer conditions in 75-100\% splits, thus yielding significantly higher scores in FID. When our method without the LSP module, it exhibits significant degradation of FID metrics on rarer texts. This indicates that under rare text conditions, LLMs Semantic Parsing can provide more beneficial prior knowledge, thereby obtaining better generalization and generating motion that matches text more effectively.

% \textcolor{red}{We explored how the fine-grained descriptions parsed by LLMs affect motion generation, as illustrated in Figure \ref{llm2-final}. The text prompt is ``a person kicks with the right leg'' and the standard motion visualization is depicted in Figure \ref{llm2-final}b. We maintain the text prompt in its original content while modifying the meaning of the fine-grained descriptions parsed by the LLMs to investigate their influence on motion generation. To be precise, when we enrich the LLMs-parsed content with terms containing the meaning of ``kick higher", the motion generated is illustrated in Figure \ref{llm2-final}a. Similarly, by integrating phrases into the LLMs-parsed content that indicate the meaning of ``kick lower", the motion generated is captured in Figure \ref{llm2-final}c. Variations in the LLM-parsed descriptions, like ``kick higher" and ``kick lower", led to subtle changes in the kick's height. This reveals that Fg-T2M++ aligns closely with the overall semantics of the text prompt, and the content parsed by LLMs exerts a subtle influence on the final motion presentation.  Additionally, the detailed content extracted by LLMs exhibits a rich diversity, which in turn enriches the variety of motion generation.}

We acknowledge that the descriptions generated by LLMs may not always be completely consistent with the GT motions, especially when the text prompts become ambiguous. However, Fg-T2M++ can still complete effective motion generation, as the fine-grained descriptions provided by current LLMs are used as reference information, not as strong constraints to limit the model.

\begin{figure}[t]
    \centering
    \includegraphics[width=\linewidth]{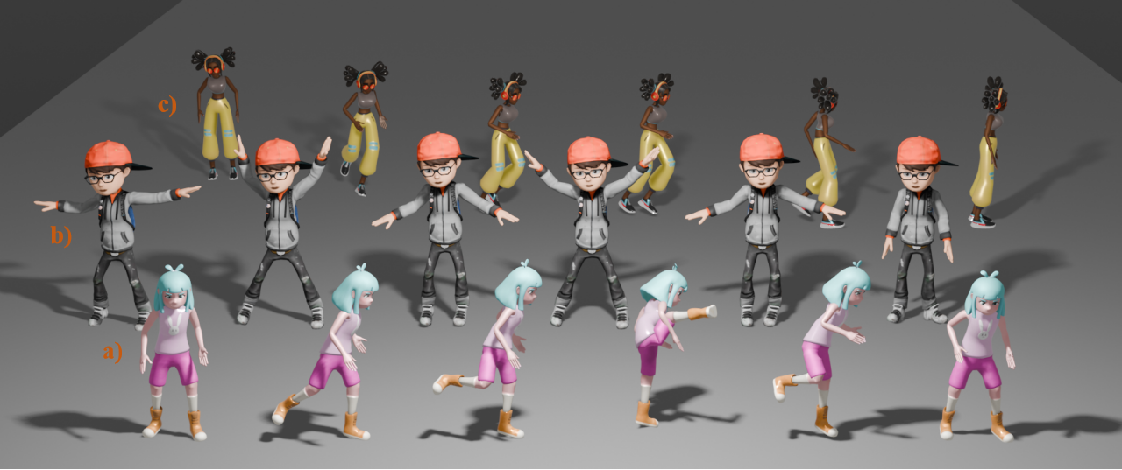}
    \caption{\textbf{Visual results on the effects of LLMs.} Motion frames are ordered from left to right.}
    \label{llm2-final}
\end{figure}

We conducted three experiments comparing the fine-grained descriptions generated by the LLMs different from the ground truth (GT), as shown in Figure \ref{llm2-final}. The first scenario is where LLMs generated descriptions that represent the same action meaning but different variants, as shown in Figure \ref{llm2-final}a. The text prompt is: ``A person kicks with the right leg''. However, when LLMs parsed the detailed action of the right leg, they described it as ``lifting the right leg towards a high place''. The result shows that the generated figure kicks the right leg relatively high. The second scenario is where LLMs capture the overall motion semantics but do not match the specific details, as shown in Figure \ref{llm2-final}b. The text prompt is: ``A person does two jumping jacks''. LLMs may focus more on the action meaning brought by the jumping jacks, neglecting the depiction of ``two'' in the fine-grained description of each joint. However, the result shows that the figure can still complete the motion of two jumping jacks. The third scenario is where LLMs describe the motion of the wrong body parts, as shown in Figure \ref{llm2-final}c. The text prompt is: ``A person runs to the left''. In the current Text2Motion dataset, the directionality is mostly used to describe the left and right of the person. Therefore, this text actually describes the person running to his left side. However, due to the lack of this prior knowledge, LLMs incorrectly describe the person as running to the left side of the screen, i.e., running to the right side of the person. The result shows that the figure still completes the action of running to his left side.

\begin{figure*}[t]
    \centering
    \includegraphics[width=\linewidth]{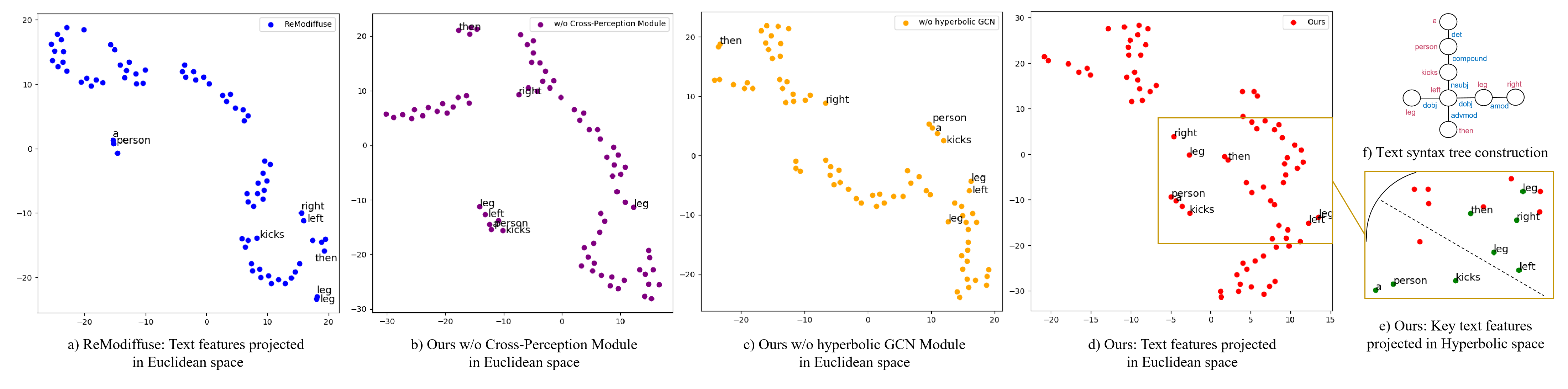}
    \caption{\textbf{Visualization of one text sample's features in hyperbolic space and Euclidean space.} a) Text feature projection of ReMoDiffuse \citep{zhang2023remodiffuse} into Euclidean space. b) Text feature projection of Fg-T2M++ without Cross-Perception Module into Euclidean space. c) Text feature projection of Fg-T2M++ without hyperbolic GCN Module into Euclidean space. d) Text feature projection of Fg-T2M++ into Euclidean space. e) Key text feature projection of Fg-T2M++ into hyperbolic space. f) Linguistic relationship tree structure of one text sample.}
    \label{treevis}
\end{figure*}

The provided examples illustrate that when the fine-grained descriptions do not align with Gt as a reference, our method can robustly generate motions that are consistent with the text prompt. Nevertheless, alleviating mismatches with GT motion is still a challenge that needs to be tackled, as more precise fine-grained parsing leads to more accurate reference information. It is possible to consider adding more task priors to the LLMs' prompts to prevent some common sense issues, such as the third scenario situation mentioned above.
 
\begin{figure}[t]
    \centering
    \includegraphics[width=0.9\linewidth]{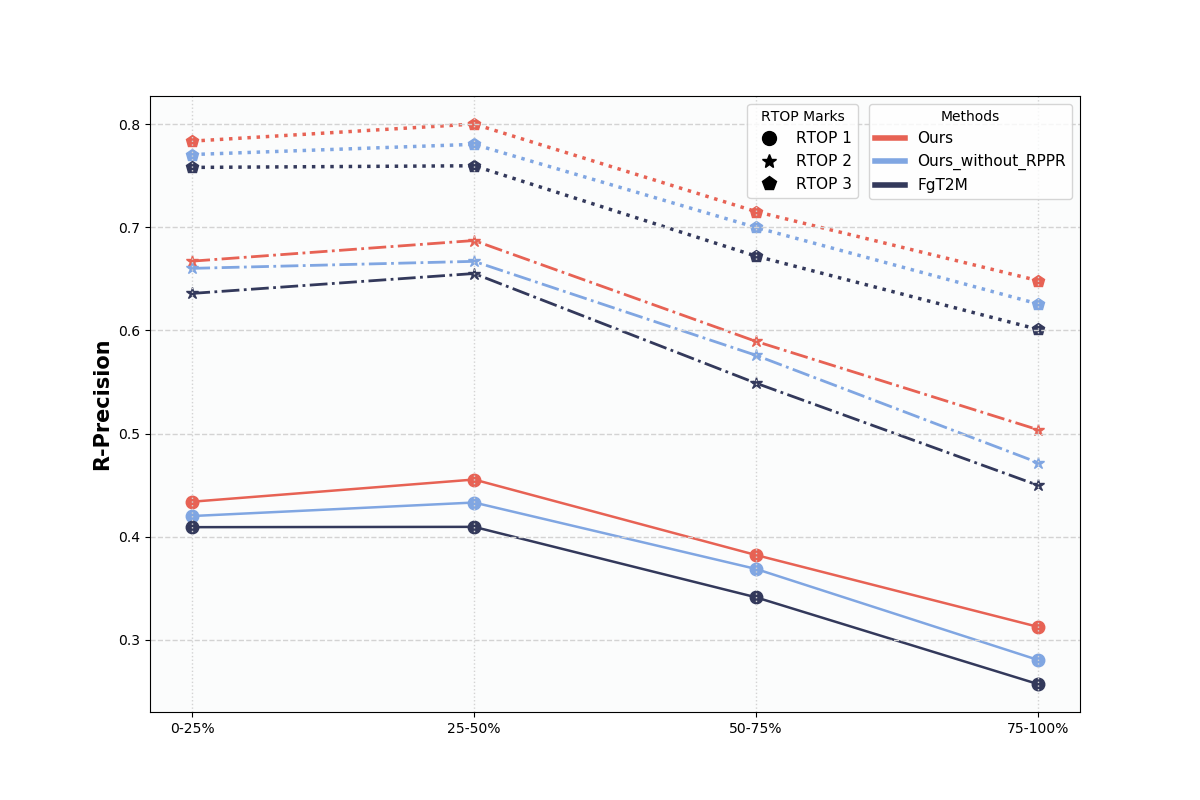}
    \caption{\textbf{Evaluation R-TOP based on different sentence lengths on KIT-ML~\citep{plappert2016kit} datasets,} where higher R-TOP indicates better performance. From 0-25\% to 75-100\% signifies increasing difficulty levels.}
    \label{R}
\end{figure}

\paragraph{The Effectiveness of the Hyperbolic Text Representation Module.} Our investigation explored the impact of textual syntactic structure and the utilization of hyperbolic space on text encoding in Table \ref{ablation}. We observed that employing a standard transformer resulted in the model struggling to capture intricate structural details within sentences, consequently leading to diminished performance. However, integrating syntactic analysis alongside graph convolutional networks, as implemented in our baseline Fg-T2M approach, significantly strengthened the model's ability to encode text, resulting in enhanced results.

A notable improvement in performance metrics—R-TOP, FID, and MM-Dists—was observed when hyperbolic GCN replaced the standard GCN. This highlights the efficacy of hyperbolic space in capturing tree structures and facilitating the seamless expansion of linguistic attributes throughout the generative process. Additionally, the introduction of a cross-perception module further refined the model's ability to assimilate fine textual nuances and LLMs-parsed features, leading to superior performance.

\begin{figure}[t]
    \centering
    \includegraphics[width=\linewidth]{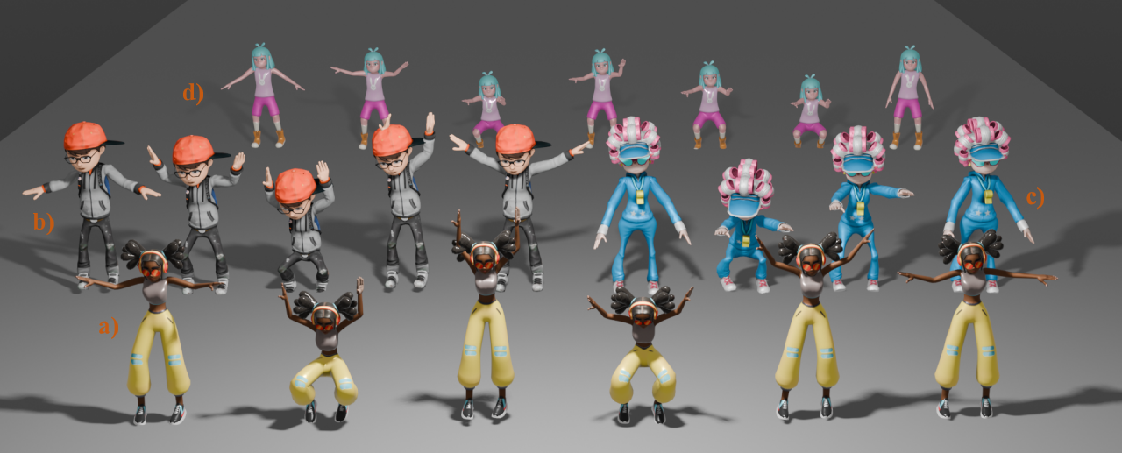}
    \caption{\textbf{Qualitative examples on the ablation study.} Motion frames are ordered from left to right. Text prompt: A person performs two squats while lifting his arms to shoulder height and hands above his head. a): Fg-T2M++. b): Fg-T2M++ w/o multi-modal fusion module. c): Fg-T2M++ w/o hyperbolic text representation module. d): Fg-T2M++ w/o LLMs semantic parsing module.}
    \label{xiaorong-final}
\end{figure}

\begin{figure*}[t]
    \centering
    \includegraphics[width=\linewidth]{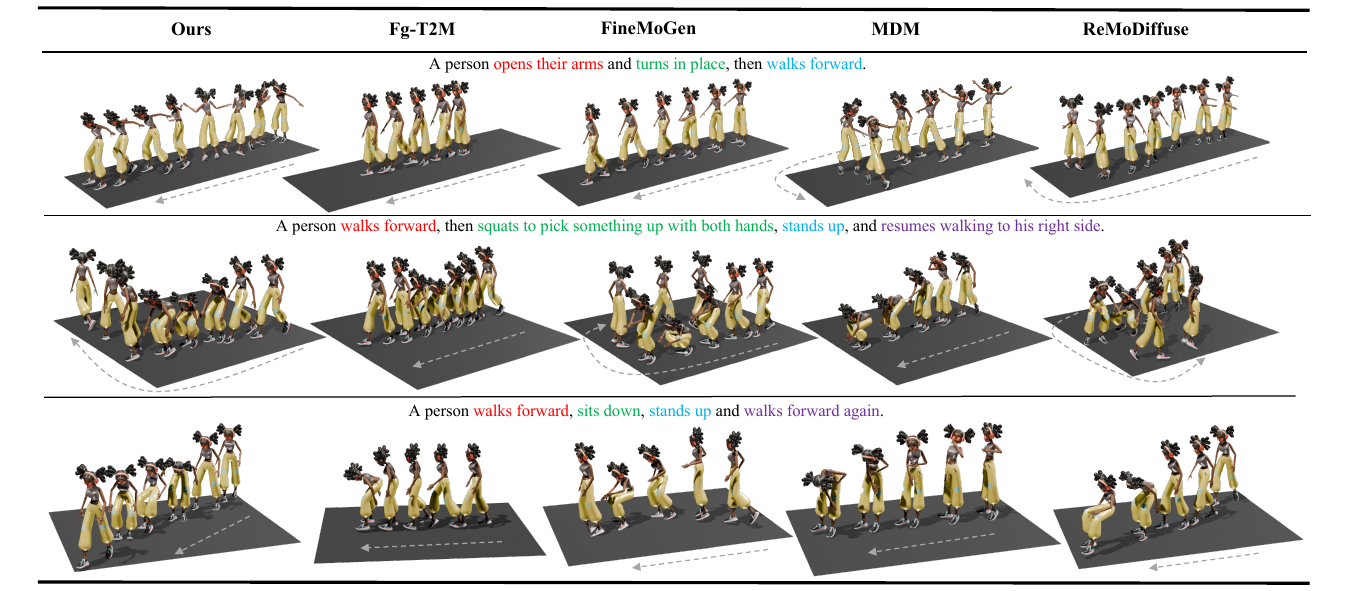}
    \caption{\textbf{Visual results compared with existing methods.} The gray arrow represents the time axes.}
    \label{compare}
\end{figure*}

The superiority of the hyperbolic text representation module is further underscored by a visualization analysis of text features in hyperbolic space, as depicted in Figure \ref{treevis}. Consider the sentence ``A person kicks left leg then right leg" for illustration. The linguistic features of the ReMoDiffuse \citep{zhang2023remodiffuse} are chaotically distributed in Euclidean space, lacking a clear tree-like hierarchical organization. Additionally, features representing similar linguistic concepts, such as ``left" and ``right," are overly condensed, as demonstrated in Figure \ref{treevis}a. In stark contrast, the linguistic features of our Fg-T2M++ are arranged more logically in Euclidean space, efficiently differentiating between similar linguistic elements, for example, ``left leg" and ``right leg," as shown in Figure \ref{treevis}d. When these text language features are projected into hyperbolic space, they present a tree-like hierarchical structure, as illustrated in Figures \ref{treevis}e and \ref{treevis}f. To verify the significance of fine-grained descriptions and the function of hyperbolic GCN within the hyperbolic text representation module, we deactivate the cross-perception and hyperbolic GCN modules to study their influence on the text feature space. As depicted in Figure \ref{treevis}b, the absence of the cross-perception module results in a compact text feature representation that may disproportionately focus on the initial action, such as "a person kicks left leg," potentially overlooking subsequent sentence elements. Furthermore, as Figure \ref{treevis}c illustrates, when the hyperbolic GCN module is removed, the text features of the left leg and right leg cannot learn a clear distinction like that in Figure \ref{treevis}d, thus posing a great challenge to the subsequent motion generation process. This showcases that Fg-T2M++, with its Hyperbolic Text Representation Module, adeptly learns the tree-like hierarchical architecture of language, thus harnessing more effective linguistic features for enhancing motion generation.

\paragraph{The Effectiveness of Multi-Modal Fusion Module.} Expanding upon our baseline Fg-T2M, which utilizes conventional word-level and sentence-level feature fusion methods, the integration of subtle, fine-grained insights from LLMs parsing at both the word and sentence levels substantially enriches the text-to-motion generation process by providing a deeper contextual understanding, as shown in Table \ref{ablation}. This strategic enhancement results in significantly improved performance. It is also observed that the fusion of word-level features, compared to sentence-level integration, has a more pronounced impact on refining and enhancing the quality of motion generation. This underscores the critical importance of linguistic analysis in advancing the fidelity of generated motions.

We further conducted additional experiments to assess the performance of the MMF Module under various lengths of text prompts. We sorted the data based on the length of the text prompts and divided it into four segments: Tail 0-25\% (less than 6 words), Tail 25-50\% (between 6 and 8 words), Tail 50-75\% (between 8 and 10 words), and Tail 75-100\% (more than 10 words), ordered from short to long. Figure \ref{R} presents the quantitative results compared with Fg-T2M \citep{wang2023fg} and Our method without the MMF module. When the text prompts are long and complex, the performance of Fg-T2M \citep{wang2023fg}, as well as our method without the MMF module, degrades significantly. However, our Fg-T2M++ shows the least degradation, demonstrating the effectiveness of our proposed MMF module, which incorporates the idea of global and local progressive fusion.

\paragraph{Motion Visualizations on the Ablation Study.} To thoroughly assess the individual contributions of each module, we employed motion visualization for an in-depth comparative analysis, as illustrated in Figure \ref{xiaorong-final}. Through ablation visualization, we examined the effects of removing key components. Upon the removal of the LLM semantic parsing module, the model can still roughly complete the overall motion but lacks detail, notably failing to raise the hands above the head as specified. Excluding the multi-modal fusion module resulted in the model's limited capability to perform a single squat. The absence of the hyperbolic text representation module led to a significant decline in performance, with inaccuracies in both the number of squats and the positioning of the hands. In contrast, our full Fg-T2M++ model adeptly executed the motions as dictated by the textual descriptions, confirming the method's ability to understand complex sentences and generate high-quality motion.

%\paragraph{Motion Visualizations on the Ablation Study.} To thoroughly assess the individual contributions of each module, we employed motion visualization for an in-depth comparative analysis. Through ablation visualization, we examined the effects of removing key components, as illustrated in Figure \ref{xiaorong-final}. Upon the removal of the LLMs semantic parsing module, the model can still roughly complete the overall motion but exhibited a lack of detail, notably failing to raise the hands above the head as specified. Excluding the multi-modal fusion module resulted in the model's limited capability to perform a single squat. The absence of the hyperbolic text representation module led to a significant decline in performance, with inaccuracies in both the number of squats and the positioning of the hands. In contrast, our Fg-T2M++ adeptly executed the motions as dictated by the textual descriptions, which confirms the method's ability to understand complex sentences and its capability to generate high-quality motion.

\begin{figure}[t]
    \centering
    \includegraphics[width=\linewidth]{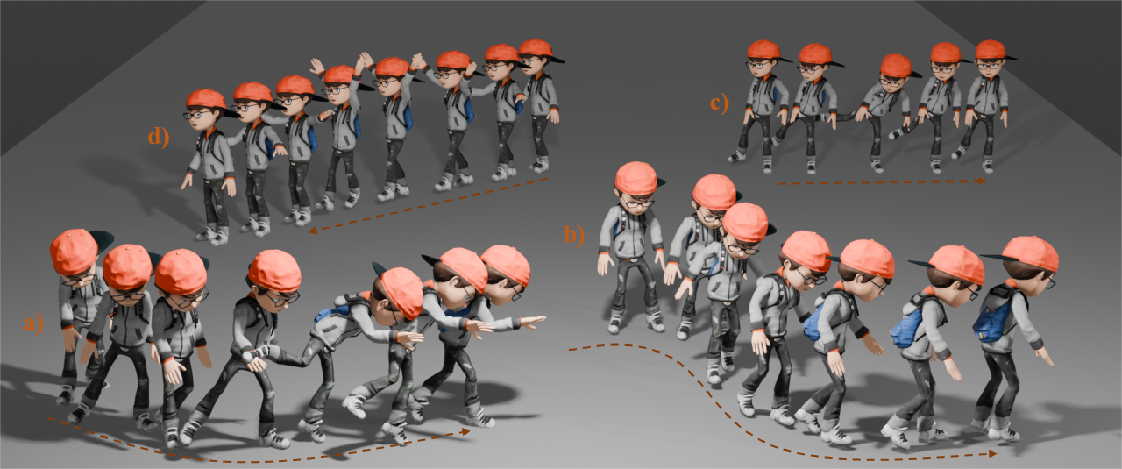}
    \caption{\textbf{More examples of visualizations.} a): A person walks forward and lifts one leg, almost tripping over something. b): A person runs in a s-shape. c): A person raises their right leg and extends it then lowers it. d): A person is walking while raising both hands.}
    \label{morevis}
\end{figure}

\subsection{Qualitative Analysis}

To highlight the effectiveness of Fg-T2M++, Figure \ref{compare} provides a qualitative comparison with Fg-T2M \citep{wang2023fg}, FineMoGen \citep{zhang2024finemogen}, MDM \citep{tevet2023human}, and ReMoDiffuse \citep{zhang2023remodiffuse}. 
By comparison, our initial version, Fg-T2M \citep{wang2023fg}, still faces challenges in capturing the intricacies within more complex sentences, often missing out on some action details.
ReMoDiffuse \citep{zhang2023remodiffuse}, employing retrieval techniques, elevates the quality of motion generation and excels across action categories but encounters challenges in generating motions that precisely align with the text descriptions. MDM \citep{tevet2023human} experiences a sharp decline in performance when faced with challenging or lengthy text prompts. FineMoGen \citep{zhang2024finemogen} captures the general essence of the text but falls short of capturing finer details. Overall, our method excels in generating high-quality motions that faithfully represent the input text, surpassing these models under complex text conditions. In Figure \ref{morevis}, we present additional visual examples, showcasing Fg-T2M++'s robust text comprehension capabilities and its proficiency in generating intricate motions.

\section{Limitations, Future Work and Conclusion}\label{sec5}

\paragraph{Limitations and Future Work.}
The effectiveness of Fg-T2M++ is closely tied to the capabilities of pre-trained large-scale language models. This reliance can present challenges, particularly in requiring applications to provide detailed and specific input formats, which to a certain extent limits its application scenarios. Additionally, the current model is limited to generating motion sequences with up to 196 frames, which restricts its application for longer sequences. Future research could focus on extending motion sequence length and achieving smooth transitions between actions to better meet real-world needs. Furthermore, modeling interactions between humans and their environments, including other people and scenes, represents another promising research direction.

\paragraph{Conclusion.}

In this paper, we introduce Fg-T2M++, a method for fine-grained text-driven human motion generation using diffusion models. Specifically, Fg-T2M++ integrates three advanced techniques: LLMs Semantic Parsing, Hyperbolic Text Representation Module, and Multi-Modal Fusion Module. By leveraging the powerful prior knowledge of LLMs to parse text prompts effectively and utilizing language relationships to construct precise language features, Fg-T2M++ achieves multi-step reasoning through hierarchical feature fusion at both global and detailed levels. Our quantitative and qualitative results demonstrate that our approach outperforms existing SOTA methods in text-driven motion generation tasks, producing high-quality, fine-grained motions that align with text prompts even under complex text conditions.

\begin{acknowledgements}
This work was supported by the National Natural Science Foundation of China (Project Number: 62272019).
\end{acknowledgements}

\paragraph{Data Availability Statement.}
In this work, we use publicly available datasets, HumanML3D and KIT. These two datasets can be obtained at \url{https://github.com/EricGuo5513/HumanML3D} and \url{https://drive.google.com/drive/folders/1MnixfyGfujSP-4t8w_2QvjtTVpEKr97t}, respectively.

\bibliographystyle{spbasic}      

\bibliography{ref}

\begin{thebibliography}{52}
\providecommand{\natexlab}[1]{#1}
\providecommand{\url}[1]{{#1}}
\providecommand{\urlprefix}{URL }
\expandafter\ifx\csname urlstyle\endcsname\relax
  \providecommand{\doi}[1]{DOI~\discretionary{}{}{}#1}\else
  \providecommand{\doi}{DOI~\discretionary{}{}{}\begingroup \urlstyle{rm}\Url}\fi
\providecommand{\eprint}[2][]{\url{#2}}

\bibitem[{Achiam et~al.(2023)Achiam, Adler, Agarwal, Ahmad, Akkaya, Aleman, Almeida, Altenschmidt, Altman, Anadkat et~al.}]{achiam2023gpt}
Achiam J, Adler S, Agarwal S, Ahmad L, Akkaya I, Aleman FL, Almeida D, Altenschmidt J, Altman S, Anadkat S, et~al. (2023) Gpt-4 technical report. arXiv preprint arXiv:230308774

\bibitem[{Ahuja and Morency(2019)}]{ahuja2019language2pose}
Ahuja C, Morency LP (2019) Language2pose: Natural language grounded pose forecasting. In: 2019 International Conference on 3D Vision (3DV), IEEE, pp 719--728

\bibitem[{Athanasiou et~al.(2023)Athanasiou, Petrovich, Black, and Varol}]{athanasiou2023sinc}
Athanasiou N, Petrovich M, Black MJ, Varol G (2023) Sinc: Spatial composition of 3d human motions for simultaneous action generation. arXiv preprint arXiv:230410417

\bibitem[{Cervantes et~al.(2022)Cervantes, Sekikawa, Sato, and Shinoda}]{cervantes2022implicit}
Cervantes P, Sekikawa Y, Sato I, Shinoda K (2022) Implicit neural representations for variable length human motion generation. In: European Conference on Computer Vision, Springer, pp 356--372

\bibitem[{Chen et~al.(2023)Chen, Jiang, Liu, Huang, Fu, Chen, and Yu}]{chen2023executing}
Chen X, Jiang B, Liu W, Huang Z, Fu B, Chen T, Yu G (2023) Executing your commands via motion diffusion in latent space. In: Proceedings of the IEEE/CVF Conference on Computer Vision and Pattern Recognition, pp 18000--18010

\bibitem[{Desai et~al.(2023)Desai, Nickel, Rajpurohit, Johnson, and Vedantam}]{desai2023hyperbolic}
Desai K, Nickel M, Rajpurohit T, Johnson J, Vedantam SR (2023) Hyperbolic image-text representations. In: International Conference on Machine Learning, PMLR, pp 7694--7731

\bibitem[{Devlin et~al.(2018)Devlin, Chang, Lee, and Toutanova}]{devlin2018bert}
Devlin J, Chang MW, Lee K, Toutanova K (2018) Bert: Pre-training of deep bidirectional transformers for language understanding. arXiv preprint arXiv:181004805

\bibitem[{Ghosh et~al.(2021)Ghosh, Cheema, Oguz, Theobalt, and Slusallek}]{ghosh2021synthesis}
Ghosh A, Cheema N, Oguz C, Theobalt C, Slusallek P (2021) Synthesis of compositional animations from textual descriptions. In: Proceedings of the IEEE/CVF international conference on computer vision, pp 1396--1406

\bibitem[{Gilardi et~al.(2023)Gilardi, Alizadeh, and Kubli}]{gilardi2023chatgpt}
Gilardi F, Alizadeh M, Kubli M (2023) Chatgpt outperforms crowd-workers for text-annotation tasks. arXiv preprint arXiv:230315056

\bibitem[{Guo et~al.(2020)Guo, Zuo, Wang, Zou, Sun, Deng, Gong, and Cheng}]{guo2020action2motion}
Guo C, Zuo X, Wang S, Zou S, Sun Q, Deng A, Gong M, Cheng L (2020) Action2motion: Conditioned generation of 3d human motions. In: Proceedings of the 28th ACM International Conference on Multimedia, pp 2021--2029

\bibitem[{Guo et~al.(2022{\natexlab{a}})Guo, Zou, Zuo, Wang, Ji, Li, and Cheng}]{guo2022generating}
Guo C, Zou S, Zuo X, Wang S, Ji W, Li X, Cheng L (2022{\natexlab{a}}) Generating diverse and natural 3d human motions from text. In: Proceedings of the IEEE/CVF Conference on Computer Vision and Pattern Recognition, pp 5152--5161

\bibitem[{Guo et~al.(2022{\natexlab{b}})Guo, Zuo, Wang, and Cheng}]{guo2022tm2t}
Guo C, Zuo X, Wang S, Cheng L (2022{\natexlab{b}}) Tm2t: Stochastic and tokenized modeling for the reciprocal generation of 3d human motions and texts. In: European Conference on Computer Vision, Springer, pp 580--597

\bibitem[{Guo et~al.(2022{\natexlab{c}})Guo, Zuo, Wang, Liu, Zou, Gong, and Cheng}]{guo2022action2video}
Guo C, Zuo X, Wang S, Liu X, Zou S, Gong M, Cheng L (2022{\natexlab{c}}) Action2video: Generating videos of human 3d actions. International Journal of Computer Vision 130(2):285--315

\bibitem[{Ho and Salimans(2022)}]{ho2022classifier}
Ho J, Salimans T (2022) Classifier-free diffusion guidance. arXiv preprint arXiv:220712598

\bibitem[{Ho et~al.(2020)Ho, Jain, and Abbeel}]{ho2020denoising}
Ho J, Jain A, Abbeel P (2020) Denoising diffusion probabilistic models. Advances in neural information processing systems 33:6840--6851

\bibitem[{Honnibal and Montani(2017)}]{honnibal2017spacy}
Honnibal M, Montani I (2017) spacy 2: Natural language understanding with bloom embeddings, convolutional neural networks and incremental parsing. To appear 7(1):411--420

\bibitem[{Jiang et~al.(2024)Jiang, Chen, Liu, Yu, Yu, and Chen}]{jiang2024motiongpt}
Jiang B, Chen X, Liu W, Yu J, Yu G, Chen T (2024) Motiongpt: Human motion as a foreign language. Advances in Neural Information Processing Systems 36

\bibitem[{Jin et~al.(2024)Jin, Wu, Fan, Sun, Yang, and Yuan}]{jin2024act}
Jin P, Wu Y, Fan Y, Sun Z, Yang W, Yuan L (2024) Act as you wish: Fine-grained control of motion diffusion model with hierarchical semantic graphs. Advances in Neural Information Processing Systems 36

\bibitem[{Kalakonda et~al.(2023)Kalakonda, Maheshwari, and Sarvadevabhatla}]{kalakonda2023action}
Kalakonda SS, Maheshwari S, Sarvadevabhatla RK (2023) Action-gpt: Leveraging large-scale language models for improved and generalized action generation. In: 2023 IEEE International Conference on Multimedia and Expo (ICME), IEEE, pp 31--36

\bibitem[{Kao and Su(2020)}]{kao2020temporally}
Kao HK, Su L (2020) Temporally guided music-to-body-movement generation. In: Proceedings of the 28th ACM International Conference on Multimedia, pp 147--155

\bibitem[{Karunratanakul et~al.(2023)Karunratanakul, Preechakul, Suwajanakorn, and Tang}]{karunratanakul2023guided}
Karunratanakul K, Preechakul K, Suwajanakorn S, Tang S (2023) Guided motion diffusion for controllable human motion synthesis. In: Proceedings of the IEEE/CVF International Conference on Computer Vision, pp 2151--2162

\bibitem[{Kim et~al.(2023)Kim, Kim, and Choi}]{kim2023flame}
Kim J, Kim J, Choi S (2023) Flame: Free-form language-based motion synthesis \& editing. In: Proceedings of the AAAI Conference on Artificial Intelligence, vol~37, pp 8255--8263

\bibitem[{Kochurov et~al.(2020)Kochurov, Karimov, and Kozlukov}]{kochurov2020geoopt}
Kochurov M, Karimov R, Kozlukov S (2020) Geoopt: Riemannian optimization in pytorch. arXiv preprint arXiv:200502819

\bibitem[{Leng et~al.(2023)Leng, Wu, Saleh, Montanaro, Yu, Wang, Navab, Liang, and Tombari}]{leng2023dynamic}
Leng Z, Wu SC, Saleh M, Montanaro A, Yu H, Wang Y, Navab N, Liang X, Tombari F (2023) Dynamic hyperbolic attention network for fine hand-object reconstruction. In: Proceedings of the IEEE/CVF International Conference on Computer Vision, pp 14894--14904

\bibitem[{Li et~al.(2021)Li, Yang, Ross, and Kanazawa}]{li2021ai}
Li R, Yang S, Ross DA, Kanazawa A (2021) Ai choreographer: Music conditioned 3d dance generation with aist++. In: Proceedings of the IEEE/CVF International Conference on Computer Vision, pp 13401--13412

\bibitem[{Liu et~al.(2019)Liu, Nickel, and Kiela}]{liu2019hyperbolic}
Liu Q, Nickel M, Kiela D (2019) Hyperbolic graph neural networks. Advances in neural information processing systems 32

\bibitem[{Loper et~al.(2023)Loper, Mahmood, Romero, Pons-Moll, and Black}]{loper2023smpl}
Loper M, Mahmood N, Romero J, Pons-Moll G, Black MJ (2023) Smpl: A skinned multi-person linear model. In: Seminal Graphics Papers: Pushing the Boundaries, Volume 2, pp 851--866

\bibitem[{Mahmood et~al.(2019)Mahmood, Ghorbani, Troje, Pons-Moll, and Black}]{mahmood2019amass}
Mahmood N, Ghorbani N, Troje NF, Pons-Moll G, Black MJ (2019) Amass: Archive of motion capture as surface shapes. In: Proceedings of the IEEE/CVF international conference on computer vision, pp 5442--5451

\bibitem[{Nickel and Kiela(2017)}]{nickel2017poincare}
Nickel M, Kiela D (2017) Poincar{\'e} embeddings for learning hierarchical representations. Advances in neural information processing systems 30

\bibitem[{Petrovich et~al.(2021)Petrovich, Black, and Varol}]{petrovich2021action}
Petrovich M, Black MJ, Varol G (2021) Action-conditioned 3d human motion synthesis with transformer vae. In: Proceedings of the IEEE/CVF International Conference on Computer Vision, pp 10985--10995

\bibitem[{Petrovich et~al.(2022)Petrovich, Black, and Varol}]{petrovich2022temos}
Petrovich M, Black MJ, Varol G (2022) Temos: Generating diverse human motions from textual descriptions. In: European Conference on Computer Vision, Springer, pp 480--497

\bibitem[{Plappert et~al.(2016)Plappert, Mandery, and Asfour}]{plappert2016kit}
Plappert M, Mandery C, Asfour T (2016) The kit motion-language dataset. Big data 4(4):236--252

\bibitem[{Punnakkal et~al.(2021)Punnakkal, Chandrasekaran, Athanasiou, Quiros-Ramirez, and Black}]{punnakkal2021babel}
Punnakkal AR, Chandrasekaran A, Athanasiou N, Quiros-Ramirez A, Black MJ (2021) Babel: Bodies, action and behavior with english labels. In: Proceedings of the IEEE/CVF Conference on Computer Vision and Pattern Recognition, pp 722--731

\bibitem[{Radford et~al.(2021)Radford, Kim, Hallacy, Ramesh, Goh, Agarwal, Sastry, Askell, Mishkin, Clark et~al.}]{radford2021learning}
Radford A, Kim JW, Hallacy C, Ramesh A, Goh G, Agarwal S, Sastry G, Askell A, Mishkin P, Clark J, et~al. (2021) Learning transferable visual models from natural language supervision. In: International conference on machine learning, PMLR, pp 8748--8763

\bibitem[{Raffel et~al.(2020)Raffel, Shazeer, Roberts, Lee, Narang, Matena, Zhou, Li, and Liu}]{raffel2020exploring}
Raffel C, Shazeer N, Roberts A, Lee K, Narang S, Matena M, Zhou Y, Li W, Liu PJ (2020) Exploring the limits of transfer learning with a unified text-to-text transformer. The Journal of Machine Learning Research 21(1):5485--5551

\bibitem[{Ren et~al.(2020)Ren, Li, Huang, and Chen}]{ren2020self}
Ren X, Li H, Huang Z, Chen Q (2020) Self-supervised dance video synthesis conditioned on music. In: Proceedings of the 28th ACM International Conference on Multimedia, pp 46--54

\bibitem[{Shafir et~al.(2023)Shafir, Tevet, Kapon, and Bermano}]{shafir2023human}
Shafir Y, Tevet G, Kapon R, Bermano AH (2023) Human motion diffusion as a generative prior. arXiv preprint arXiv:230301418

\bibitem[{Shen et~al.(2021)Shen, Zhang, Zhao, Yi, and Li}]{Shen_2021_WACV}
Shen Z, Zhang M, Zhao H, Yi S, Li H (2021) Efficient attention: Attention with linear complexities. In: Proceedings of the IEEE/CVF Winter Conference on Applications of Computer Vision (WACV), pp 3531--3539

\bibitem[{Starke et~al.(2022)Starke, Mason, and Komura}]{starke2022deepphase}
Starke S, Mason I, Komura T (2022) Deepphase: Periodic autoencoders for learning motion phase manifolds. ACM Transactions on Graphics (TOG) 41(4):1--13

\bibitem[{Terlemez et~al.(2014)Terlemez, Ulbrich, Mandery, Do, Vahrenkamp, and Asfour}]{terlemez2014master}
Terlemez {\"O}, Ulbrich S, Mandery C, Do M, Vahrenkamp N, Asfour T (2014) Master motor map (mmm)—framework and toolkit for capturing, representing, and reproducing human motion on humanoid robots. In: 2014 IEEE-RAS International Conference on Humanoid Robots, IEEE, pp 894--901

\bibitem[{Tevet et~al.(2022)Tevet, Gordon, Hertz, Bermano, and Cohen-Or}]{tevet2022motionclip}
Tevet G, Gordon B, Hertz A, Bermano AH, Cohen-Or D (2022) Motionclip: Exposing human motion generation to clip space. In: European Conference on Computer Vision, Springer, pp 358--374

\bibitem[{Tevet et~al.(2023)Tevet, Raab, Gordon, Shafir, Cohen-or, and Bermano}]{tevet2023human}
Tevet G, Raab S, Gordon B, Shafir Y, Cohen-or D, Bermano AH (2023) Human motion diffusion model. In: The Eleventh International Conference on Learning Representations, \urlprefix\url{https://openreview.net/forum?id=SJ1kSyO2jwu}

\bibitem[{Tseng et~al.(2023)Tseng, Castellon, and Liu}]{tseng2023edge}
Tseng J, Castellon R, Liu K (2023) Edge: Editable dance generation from music. In: Proceedings of the IEEE/CVF Conference on Computer Vision and Pattern Recognition, pp 448--458

\bibitem[{Vaswani et~al.(2017)Vaswani, Shazeer, Parmar, Uszkoreit, Jones, Gomez, Kaiser, and Polosukhin}]{vaswani2017attention}
Vaswani A, Shazeer N, Parmar N, Uszkoreit J, Jones L, Gomez AN, Kaiser {\L}, Polosukhin I (2017) Attention is all you need. Advances in neural information processing systems 30

\bibitem[{Wan et~al.(2023)Wan, Dou, Komura, Wang, Jayaraman, and Liu}]{wan2023tlcontrol}
Wan W, Dou Z, Komura T, Wang W, Jayaraman D, Liu L (2023) Tlcontrol: Trajectory and language control for human motion synthesis. arXiv preprint arXiv:231117135

\bibitem[{Wang et~al.(2023)Wang, Leng, Li, Wu, and Liang}]{wang2023fg}
Wang Y, Leng Z, Li FW, Wu SC, Liang X (2023) Fg-t2m: Fine-grained text-driven human motion generation via diffusion model. In: Proceedings of the IEEE/CVF International Conference on Computer Vision, pp 22035--22044

\bibitem[{Yang et~al.(2022)Yang, Zhou, Li, Liu, Pan, Xiong, and King}]{yang2022hyperbolic}
Yang M, Zhou M, Li Z, Liu J, Pan L, Xiong H, King I (2022) Hyperbolic graph neural networks: a review of methods and applications. arXiv preprint arXiv:220213852

\bibitem[{Zhang et~al.(2023{\natexlab{a}})Zhang, Zhang, Cun, Zhang, Zhao, Lu, Shen, and Shan}]{zhang2023generating}
Zhang J, Zhang Y, Cun X, Zhang Y, Zhao H, Lu H, Shen X, Shan Y (2023{\natexlab{a}}) Generating human motion from textual descriptions with discrete representations. In: Proceedings of the IEEE/CVF Conference on Computer Vision and Pattern Recognition, pp 14730--14740

\bibitem[{Zhang et~al.(2023{\natexlab{b}})Zhang, Guo, Pan, Cai, Hong, Li, Yang, and Liu}]{zhang2023remodiffuse}
Zhang M, Guo X, Pan L, Cai Z, Hong F, Li H, Yang L, Liu Z (2023{\natexlab{b}}) Remodiffuse: Retrieval-augmented motion diffusion model. arXiv preprint arXiv:230401116

\bibitem[{Zhang et~al.(2024{\natexlab{a}})Zhang, Cai, Pan, Hong, Guo, Yang, and Liu}]{zhang2022motiondiffuse}
Zhang M, Cai Z, Pan L, Hong F, Guo X, Yang L, Liu Z (2024{\natexlab{a}}) Motiondiffuse: Text-driven human motion generation with diffusion model. IEEE Transactions on Pattern Analysis and Machine Intelligence pp 1--15, \doi{10.1109/TPAMI.2024.3355414}

\bibitem[{Zhang et~al.(2024{\natexlab{b}})Zhang, Li, Cai, Ren, Yang, and Liu}]{zhang2024finemogen}
Zhang M, Li H, Cai Z, Ren J, Yang L, Liu Z (2024{\natexlab{b}}) Finemogen: Fine-grained spatio-temporal motion generation and editing. Advances in Neural Information Processing Systems 36

\bibitem[{Zhong et~al.(2023)Zhong, Hu, Zhang, and Xia}]{zhong2023attt2m}
Zhong C, Hu L, Zhang Z, Xia S (2023) Attt2m: Text-driven human motion generation with multi-perspective attention mechanism. In: Proceedings of the IEEE/CVF International Conference on Computer Vision, pp 509--519

\end{thebibliography}

\clearpage
\appendix
\section*{Appendix}
\section{User Study}
\begin{figure}[h]
    \centering
    \includegraphics[width=\linewidth]{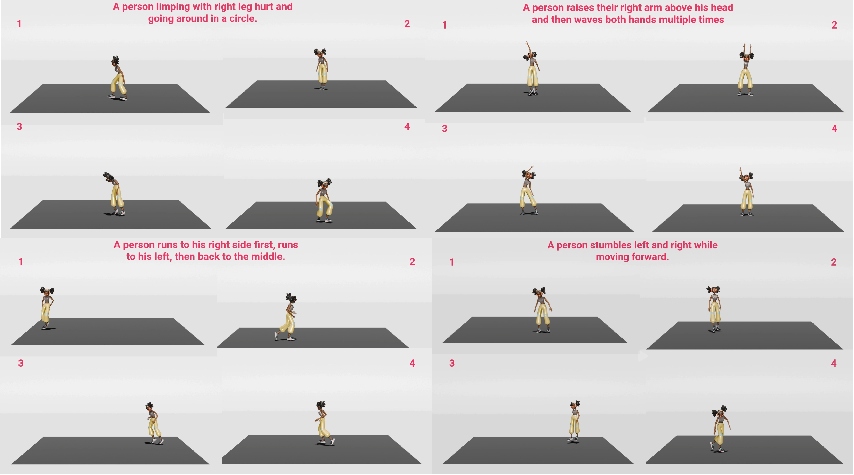}
    \caption{Visualization comparison page of our user study. }
    \label{userstudy_page}
\end{figure}

Figure \ref{userstudy_page} shows the comparison page for our user study. For each text prompt, the motion is generated through different methods and randomly reshuffled. Users are required to rank their preferences for the given motions. The motions in the user study include the category of fine-grained motions, such as ``a person is limping with the right leg hurt and going around in a circle'' or ``a person is raising his right arms above his head and then waves both hands multiple times.'' These text prompts contain many detailed features, requiring the generation methods to fully capture and model fine-grained features. The user study also includes the category of long sequence motions, such as ``a person walks forward, then squats to pick something up with both hands, stands up, and resumes walking to his right side'' or ``a person walks forward, sits down, stands up, and walks forward again.'' These text prompts contain complex combinations of multiple actions, challenging the model's ability to learn and comprehend long sequence features from text prompts.

\section{Failure Cases}

\begin{figure}[t]
    \centering
    \includegraphics[width=\linewidth]{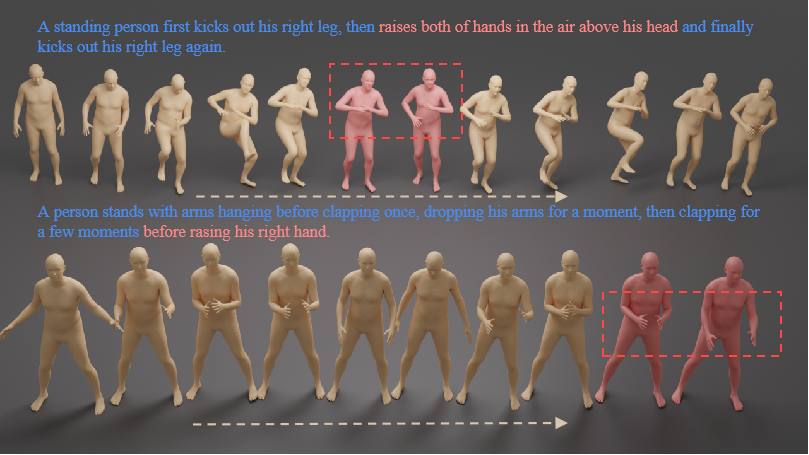}
    \caption{Visualization of some failure cases. The arrow represents the time axes and the red box indicates the incorrect motion frames.}
    \label{failure}
\end{figure}

We acknowledge some limitations in our approach, as depicted in Figure \ref{failure}. While Fg-T2M++ demonstrates proficiency in capturing the subtle details embedded in text prompts, it encounters challenges when processing lengthy sentences. This occasionally leads to the omission of certain specific actions and, consequently, results in suboptimal outcomes. To address this limitation, we propose a potential solution: strategically dividing longer sentences into multiple distinct tasks for independent processing. This approach could facilitate the preservation of fine-grained characteristics in the generation of long sequence actions.

\section{More Analysis of Text Features}

\begin{figure*}[h]
    \centering
    \includegraphics[width=\linewidth]{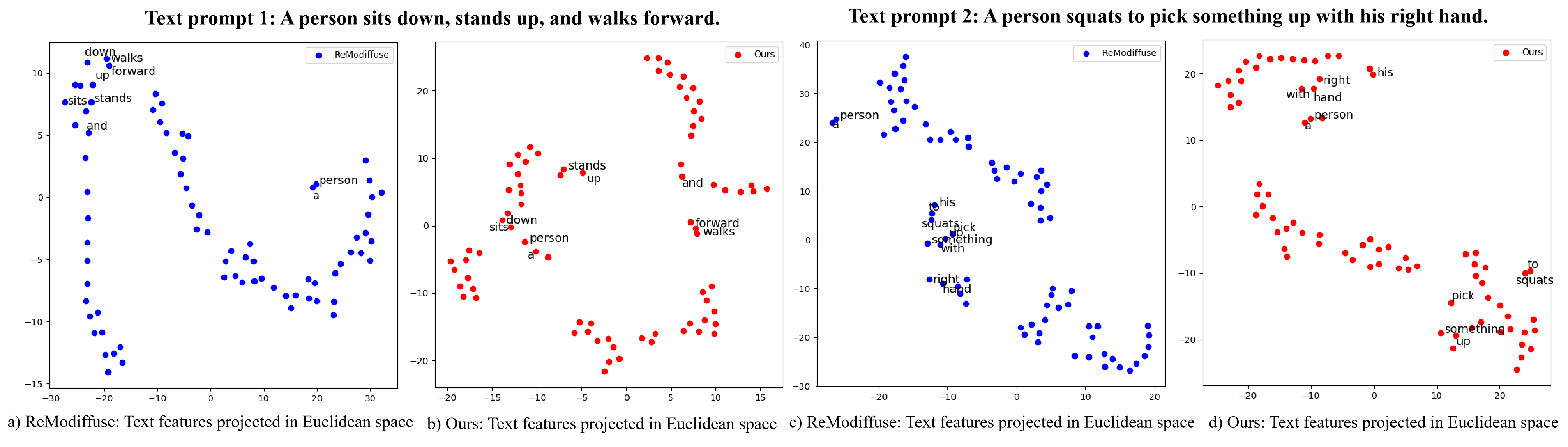}
    \caption{\textbf{Visualization of more text samples' features in hyperbolic space and Euclidean space.} a) Text 1 feature projection of ReMoDiffuse \citep{zhang2023remodiffuse} into Euclidean space. b) Text 1 feature projection of Fg-T2M++ into Euclidean space. c) Text 2 feature projection of ReMoDiffuse \citep{zhang2023remodiffuse} into Euclidean space. d) Text 2 feature projection of Fg-T2M++ into Euclidean space.}
    \label{moretextfeature}
\end{figure*}

We present further visualization analysis of text features, as shown in Figure \ref{moretextfeature}. The figure takes the sentence ``A person sits down, stands up, and walks forward'' as an example. The text features learned by ReMoDiffuse \citep{zhang2023remodiffuse} confine a series of actions to a compact space, failing to effectively distinguish the differences between each motion. As demonstrated in the third row of Figure \ref{compare} for the ReMoDiffuse \citep{zhang2023remodiffuse} case, it only manages to complete the action of sitting down, thereby neglecting the detailed features of the subsequent series of motions. In contrast, Fg-T2M++ learns the differences between each motion, effectively distinguishing the execution of different motions corresponding to different texts. 
Likewise, in the scenario depicted by text prompt 2, ``a person squats to pick something up with his right hand," ReMoDiffuse \citep{zhang2023remodiffuse} to learning textual features within a narrow text space, indicating potential overfitting during the model's training phase, which compromises its capacity for generalization. Conversely, Fg-T2M++ is adept at discerning clear and significant textual features, which in turn creates more impactful conditions that enhance the subsequent generation of motion. Therefore, it can generate high-fidelity, high-quality motion more effectively.

To verify the advantages of hyperbolic space on quantitative experiments, we calculated the geodesic distance in hyperbolic space for ReMoDiffuse \citep{zhang2023remodiffuse} and Fg-T2M++. As shown in Table \ref{order}, $D_1$, $D_2$, and $D_3$ represent the average distance from the first, second, and third layer nodes to the root node in the text-tree structure, respectively. We found that the order calculated by ReMoDiffuse \citep{zhang2023remodiffuse} is $D_2 < D_3 < D_1$, which does not conform to the hierarchical structure of the tree. In contrast, the order of Fg-t2m++ is $D_1 < D_2 < D_3$, correctly reflecting the hierarchy from the first to the third layer, revealing that our method better preserves the text-tree hierarchy.

% \begin{table}[h]
% \centering
% \resizebox{0.9\columnwidth}{!}{%
% \begin{tabular}{@{}lccccc@{}}
% \toprule \multirow{2}{*}{Method}  & \multirow{2}{*}{D_1  $\downarrow$} &\multirow{2}{*}{D_2$\downarrow$} & \multirow{2}{*}{D_3$\downarrow$}& \multirow{2}{*}{Order} \\
%   \\  \midrule
% ReMoDiffuse & $12.25$  & $11.07$  & $11.96$ & $D_2 < D_3 < D_1$\\
% Fg-T2M++ & $\textbf{4.01}$  & $\textbf{4.30}$  & $\textbf{4.45}$& $D_1 < D_2 < D_3$\\
% \bottomrule
% \end{tabular}}
% \caption{Comparing the performance in maintaining the hierarchical structure}
% \label{order}
% \end{table}

\begin{table}[h]
\centering
\resizebox{0.9\columnwidth}{!}{%
\begin{tabular}{@{}lcccc@{}}
\toprule
\multirow{2}{*}{Method} & \multirow{2}{*}{$D_1 \downarrow$} & \multirow{2}{*}{$D_2 \downarrow$} & \multirow{2}{*}{$D_3 \downarrow$} & \multirow{2}{*}{Order} \\
\cmidrule{1-5} \\
\midrule
ReMoDiffuse & $12.25$ & $11.07$ & $11.96$ & $D_2 < D_3 < D_1$ \\
Fg-T2M++ & $\textbf{4.01}$ & $\textbf{4.30}$ & $\textbf{4.45}$ & $D_1 < D_2 < D_3$ \\
\bottomrule
\end{tabular}}
\caption{Comparing the performance in maintaining the hierarchical structure}
\label{order}
\end{table}

% \textcolor{blue}{Thirdly, we presented the feature alignment between the text and the motion, as shown in Figure \ref{textmotion}. Compared to ReMoDiffuse \cite{zhang2023remodiffuse}, our method achieves closer alignment between the text-motion feature spaces, resulting in better cross-modal feature alignment.}

\begin{figure}[h]
    \centering
    \includegraphics[width=\linewidth]{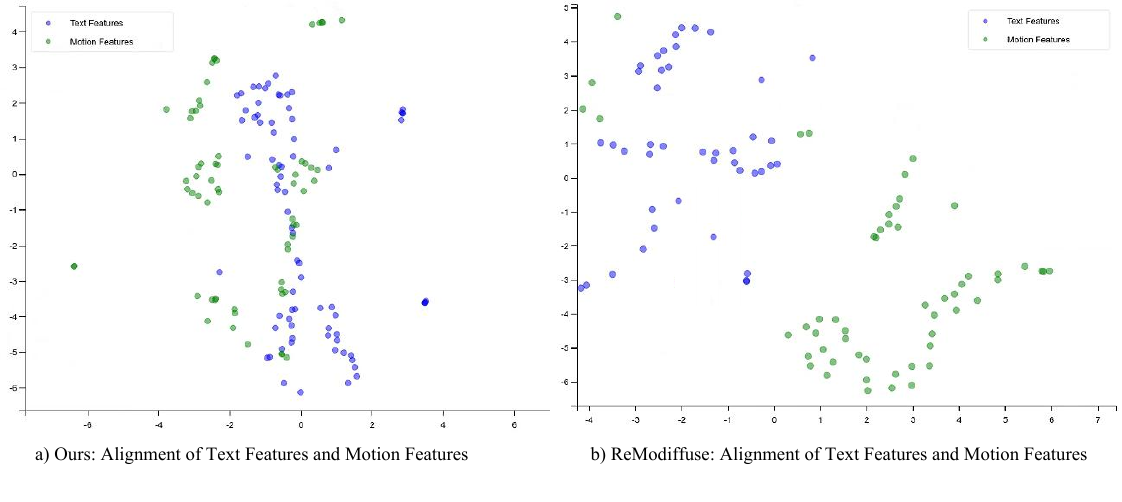}
    \caption{The text-motion feature alignment.}
    \label{textmotion}
\end{figure}

Thirdly, we demonstrated the alignment between the text feature space and the motion feature space, as shown in Figure \ref{textmotion}. Our method achieves closer alignment between these spaces compared to ReMoDiffuse \citep{zhang2023remodiffuse}, resulting in improved cross-modal feature alignment.

Moreover, the scalability and effectiveness of hyperbolic representations over transformers have been validated in large-scale settings by works such as MERU \citep{desai2023hyperbolic}, which leverage hyperbolic geometry to better preserve hierarchical relationships.

\section{Visual Comparison against Different Methods}

\begin{figure*}[t]
    \centering
    \includegraphics[width=\linewidth]{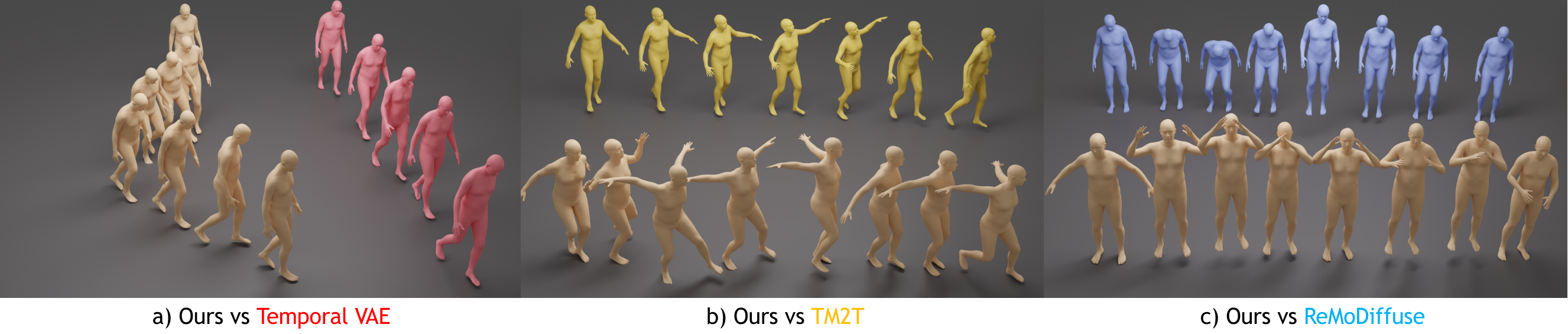}
    \caption{\textbf{Visualization comparison with different methods.} a) Compare with latent space alignment method, \textcolor{red}{Temporal VAE in red motion}, under text ``a person is walking forward while stumbling to the left.'' b) Compare with autoregressive model method, \textcolor{yellow}{TM2T in yellow motion}, under text ``a person is walking with arms swinging.'' c) Compare with diffusion model method,  \textcolor{blue}{ReMoDiffuse in blue motion}, under text ``a person is jumping while clapping.''}
    \label{related_fig}
\end{figure*}

We highlighted the limitations of other classes of methods when dealing with certain categories of sentences in the related work section. In this section, we provide a detailed demonstration to validate that Fg-T2M++ is capable of addressing these issues. As illustrated in Figure \ref{related_fig}, the motions generated by our method are depicted in cool white across three images. Figure \ref{related_fig}a demonstrates a comparison between Fg-T2M++ and the latent space alignment method, Temporal VAE \citep{guo2022generating}, revealing that Temporal VAE only produces forward-walking motion while neglecting to stumble to the left. Figure \ref{related_fig}b shows a comparison between Fg-T2M++ and the autoregressive method TM2T \citep{guo2022tm2t}, which erroneously only performs a single hand swing. Figure \ref{related_fig}c presents a comparison between Fg-T2M++ and the diffusion model method ReMoDiffuse \citep{zhang2023remodiffuse}, indicating that ReMoDiffuse fails to depict jumping and clapping simultaneously, completing only a single action. In contrast, Fg-T2M++ comprehensively generates motions consistent with the text prompts, demonstrating its superiority in handling fine-grained details.

\begin{figure}[t]
    \centering
    \includegraphics[width=\linewidth]{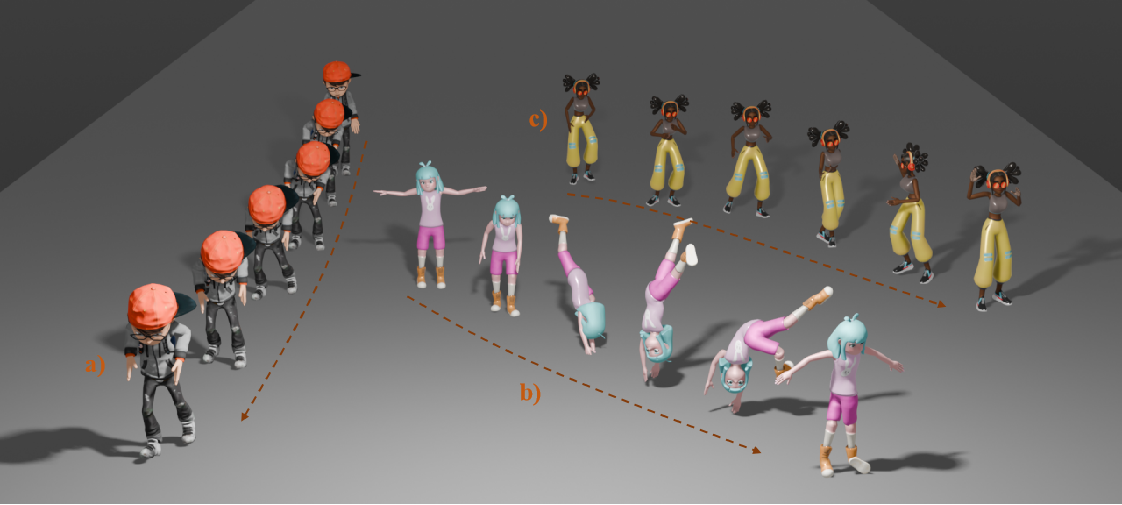}
    \caption{More diverse qualitative samples. The arrow represents the time axes. a): A person jogs forward and looks at the ground. b): A person does a cart wheel. c): A person appears to dance. }
    \label{morevis1-final}
\end{figure}

\begin{figure}[t]
    \centering
    \includegraphics[width=\linewidth]{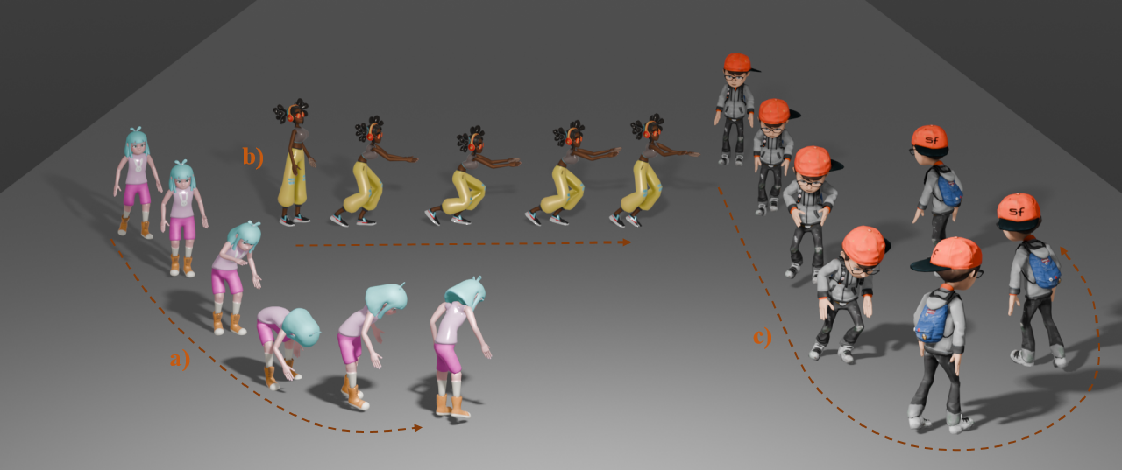}
    \caption{More diverse qualitative samples. a): A person walks to the right and picks something up. b): A person is performing lunges. c): A person runs forward and jumps over something, then turns around. }
    \label{morevis2-final}
\end{figure}

\section{More Diverse examples}

We present additional visualization examples, as shown in Figures \ref{morevis1-final} and \ref{morevis2-final}. These demonstrate Fg-T2M++'s ability to understand complex motion descriptions and its capability to generate high-fidelity, high-quality human motion.

\section{Dependency Analysis of Fg-T2M++ on LLMs}
\begin{figure}[t]
    \centering
    \includegraphics[width=\linewidth]{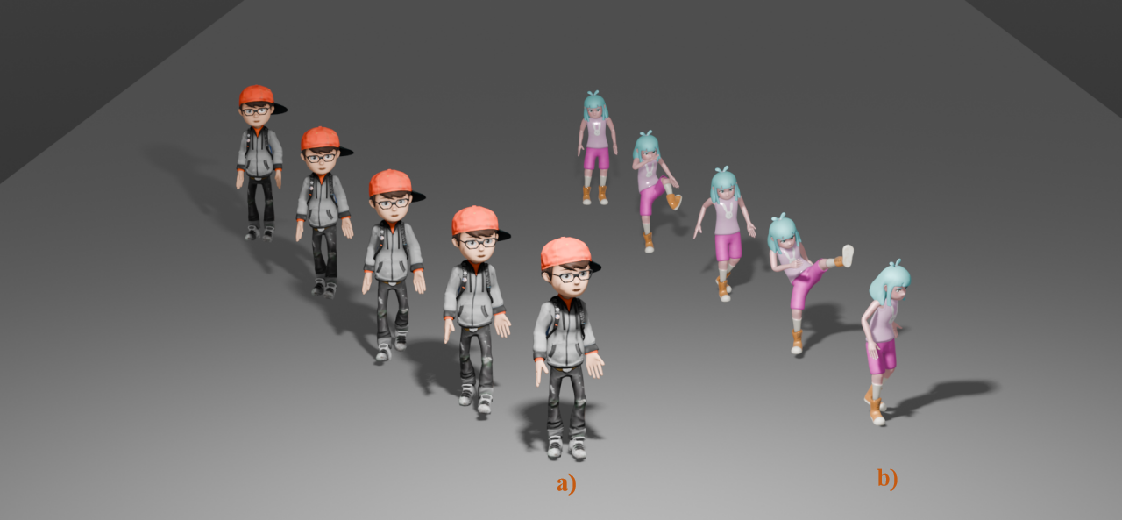}
    \caption{Dependency analysis on LLMs}
    \label{llm-final1}
\end{figure}

We discuss the scenario where LLMs provide coarse-grained text descriptions. As shown in Figure \ref{llm-final1}a, our original text prompt is: ``A person takes three steps forward.'' However, the coarse-grained content parsed by the LLMs lacks the number of steps. Similarly, the text prompt for Figure \ref{llm-final1}b is: ``A person kicks the left leg twice," while the coarse-grained content parsed by the LLMs lacks the content of ``twice." Fg-T2M++ can still generate the motion of walking and kicking, and can accurately complete the fine-grained requirement of three steps and twice. This verifies the robustness of Fg-T2M++ when the performance of LLMs is not satisfactory.

Through the analysis of the experiments, we found that the generation capability of Fg-T2M++ is not significantly affected by LLMs, i.e. when LLMs perform poorly, Fg-T2M++ does not perform poorly either. This is because Fg-T2M++ is actually more in line with the semantics of the text prompt as a whole, as the fine-grained descriptions provided by current LLMs are more used as reference information, not as strong constraints to limit the model.

\section{Performance differences between GPT-3.5 and GPT-4.}
To assess the performance differences between GPT-3.5 and GPT-4, we conducted two types of evaluations. First, in our quantitative evaluation, we randomly sampled 100 examples from the HumanML3D dataset \citep{guo2022generating} and processed the text with both GPT-3.5 and GPT-4. We evaluated the motion generation quality using metrics such as R-TOP, FID, and MultiModal Dist, as shown in Table \ref{gpt34-1}. The results indicated that replacing the LLMs parsing module with GPT-4 led to improvements in these metrics. This enhancement is attributed to GPT-4's ability to provide more detailed and comprehensive text analysis, which is beneficial for subsequent text feature extraction.

% To evaluate the performance differences between GPT-3.5 and GPT-4, we employed two experiments. The first is quantitative evaluation. We randomly sampled 100 examples from the HumanML3D dataset \citep{guo2022generating} and processed the text with both GPT-3.5 and GPT-4. We then quantitatively assessed the differences in motion generation quality, as shown in Table \ref{gpt34-1}. We found that when we replaced the large language model's parsing module with GPT-4, the model achieved performance improvements in R-TOP, FID, and MultiModal Dist, which are metrics for evaluating the quality of motion generation. This can be attributed to GPT-4's more detailed and comprehensive analysis of text, which holds greater value for subsequent text feature extraction.

\begin{table}[h]
\centering
\resizebox{0.9\columnwidth}{!}{%
\begin{tabular}{@{}lccccc@{}}
\toprule \multirow{2}{*}{Method}  & \multirow{2}{*}{R-TOP3  $\uparrow$} &\multirow{2}{*}{FID $\downarrow$} & \multirow{2}{*}{MultiModal Dist $\downarrow$}\\
  \\  \midrule
Ours with GPT-3.5 & $0.75$  & $0.73$  & $3.26$ \\
Ours with GPT-4 & $\textbf{0.77}$  & $\textbf{0.68}$  & $\textbf{3.15}$\\
\bottomrule
\end{tabular}%
}
\caption{The quantitative performance differences between GPT-3.5 and GPT-4.}
\label{gpt34-1}
\end{table}

\begin{figure}[h]
    \centering
    \includegraphics[width=\linewidth]{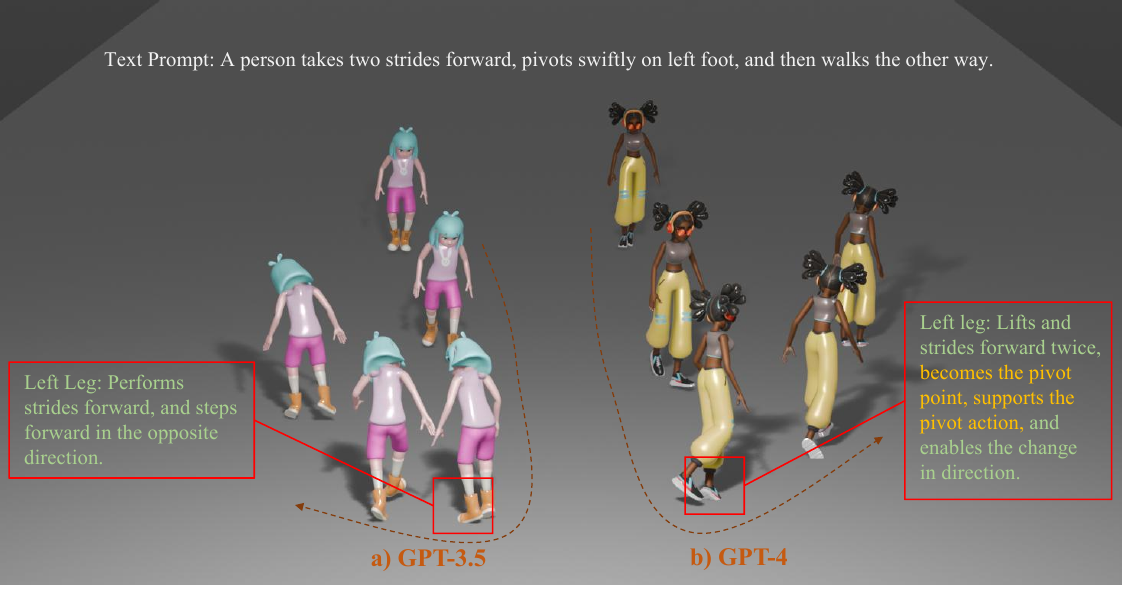}
    \caption{The visualization performance differences between GPT-3.5 and GPT-4.}
    \label{gpt34-2}
\end{figure}

Second, for qualitative evaluation, we visualized the motions generated from text prompts processed by both GPT-3.5 and GPT-4. As illustrated in Figure \ref{gpt34-2}, we highlighted differences in parsing specific details like ``left leg." GPT-3.5 failed to parse the detail of using the left foot as the root for pivot action, incorrectly using the ``right foot" instead. In contrast, GPT-4's superior parsing capability allowed for a fine-grained analysis of the left foot supporting pivot action, resulting in a motion sequence that accurately matched the text description.

% To more precisely discern the performance differences between GPT-3.5 and GPT-4, we conducted the second experiment on qualitative evaluation. We processed the text prompt through GPT-3.5 and GPT-4, and visualized the resulting motions. As shown in Figure \ref{gpt34-2}, we highlighted the differences in parsing the ``left leg'' between GPT-3.5 and GPT-4. We found that GPT-3.5 did not parse the detail of using the left foot as the root for pivot action, thus adopting the incorrect ``right foot" in the motion generation process. In contrast, GPT-4's parsing capability is powerful, with fine-grained analysis of the left foot supporting pivot action, resulting in a motion sequence that matches the text description.}

\section{LLM-parsed Fine-grained Descriptions}
\begin{figure}[h]
    \centering
    \includegraphics[width=\linewidth]{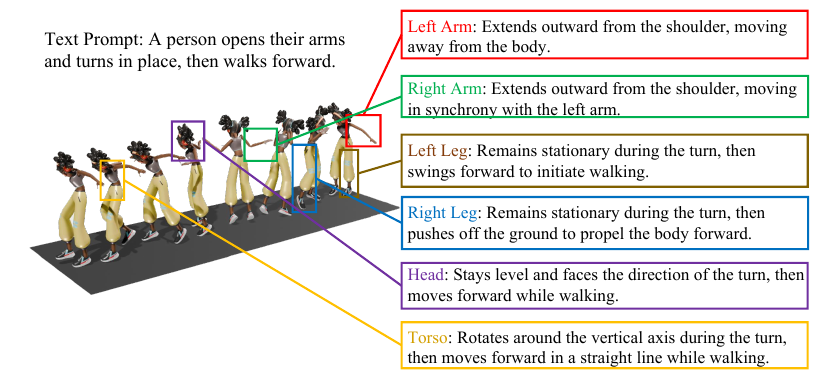}
    \caption{Correspondence between body joints and visual motion.}
    \label{Correspondence}
\end{figure}

\begin{figure*}[t]
    \centering
    \includegraphics[width=\linewidth]{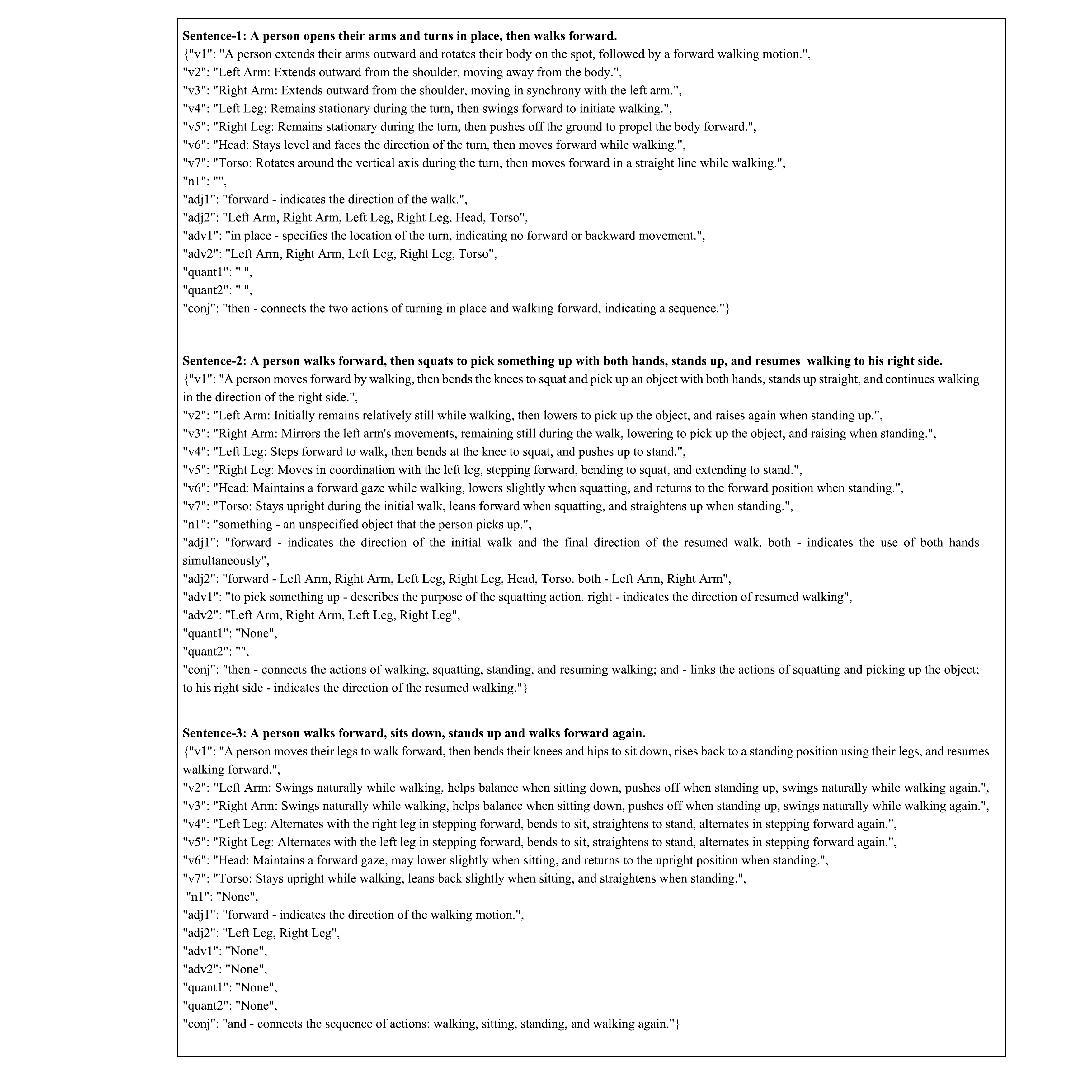}
    \caption{LLM-parsed fine-grained descriptions.}
    \label{compare_gpt}
\end{figure*}

In this section, we present some detailed content parsed by LLMs as shown in Figure \ref{compare_gpt}. These three sentences are derived from the examples in Figure \ref{compare}. It can be seen that LLMs parse the original sentence into fine-grained actions for each joint part. To be more specific, take the first sentence as an example. 
We presented text prompts, LLM-parsed fine-grained descriptions, and corresponding motion visualizations, as illustrated in Figure \ref{Correspondence}. The parsing by LLMs of parts such as the left leg, right leg, and left arm is accurately represented in the visualizations, demonstrating enhancements in connecting text descriptions to body parts. Meanwhile, as demonstrated in the top row of Figure \ref{compare}, Fg-T2M++ was the sole method to successfully achieve this precise movement. However, other approaches did not accurately reflect the arm positioning as dictated by the text description. Fg-T2M++ generates high-quality motion that conforms to the fine-grained textual description based on the LLMs-parsed output.

\end{document}